%% file: main_arxiv.tex
\documentclass[10pt, twocolumn]{article}
\usepackage[letterpaper,margin=1in,footskip=0.25in]{geometry}
\usepackage{microtype}
\usepackage[T1]{fontenc}
\usepackage[utf8]{inputenc}
\usepackage{stix2}
\usepackage{mathtools, amssymb, amsthm}
\usepackage[shortlabels]{enumitem}
\setlist[enumerate]{label=\((\roman*)\),ref=(\roman*)}
\setlist{nosep}
\usepackage{natbib}
\usepackage{hyperref}
\usepackage[nameinlink]{cleveref}
\hypersetup{
  colorlinks=true,
  linkcolor={blue!50!black},
  citecolor={green!50!black},
  urlcolor={magenta!80!black},
  bookmarks=true,
  bookmarksdepth=2,
  pdftitle = {Model-in-the-loop evaluation of retrieval systems},
  pdfauthor = {Charles Godfrey, Ping Nie and Souheil Inati}
}
\usepackage{graphicx}
\usepackage{subcaption}
\usepackage{booktabs}
\usepackage{multirow}
\usepackage{listings}
\usepackage{xcolor}
\usepackage{placeins}
\usepackage{afterpage}
\usepackage{algorithm}
\usepackage{algorithmic}

\lstdefinestyle{commoncode}{
  basicstyle=\ttfamily\footnotesize,
  breaklines=true,
  columns=fullflexible,
  tabsize=4
}

\lstdefinestyle{pythoncode}{
  style=commoncode,
  language=Python
}

\lstnewenvironment{plaincode}{
  \lstset{
    style=commoncode
  }
}{}

\lstnewenvironment{pythoncode}{
  \lstset{
    style=pythoncode
  }
}{}

\theoremstyle{plain}

\theoremstyle{definition}

\theoremstyle{remark}

\title{Likert or Not:\\\Large{LLM Absolute Relevance Judgments on Fine-Grained Ordinal Scales}}
\author{Charles Godfrey, Ping Nie, Natalia Ostapuk, David Ken, Shang Gao and Souheil Inati\\
    \texttt{\{first\}.\{last\}@thomsonreuters.com}\\
}
\date{}

\begin{document}
\maketitle

\abstract{
    \input{abstract.tex}
}

\input{body.tex}

\section{Acknowledgments}
Thanks to Isabelle Moulinier for guidance on benchmark dataset creation and
evaluation. Thanks to David Hershey for numerous helpful conversations on LLM
ranking early in the experimental design phase.

\afterpage{\FloatBarrier}

\bibliographystyle{abbrvnat}
\bibliography{main}

\clearpage
\onecolumn
\appendix

\input{appendix.tex}

\end{document}

%% file: abstract.tex
Large language models (LLMs) obtain state of the art zero shot relevance
ranking performance on a variety of information retrieval tasks. The two
most common prompts to elicit LLM relevance judgments are pointwise scoring
(a.k.a. relevance generation), where the LLM sees a single
query-document pair and outputs a single relevance score, and listwise
ranking (a.k.a. permutation generation), where the LLM sees a query
and a list of documents and outputs a permutation, sorting the documents in
decreasing order of relevance. The current research community consensus is that
listwise ranking yields superior performance, and significant research
effort has been devoted to crafting LLM listwise ranking algorithms. The
underlying hypothesis is that LLMs are better at making relative relevance
judgments than absolute ones. In tension with this hypothesis, we find that
the gap between pointwise scoring and listwise ranking shrinks when pointwise
scoring is implemented using a sufficiently large ordinal relevance label
space, becoming statistically insignificant for many LLM-benchmark dataset
combinations (where ``significant'' means ``95\% confidence that listwise
ranking improves NDCG@10''). Our evaluations span four LLMs, eight benchmark
datasets from the BEIR and TREC-DL suites, and two proprietary datasets with
relevance labels collected after the training cut-off of all LLMs
evaluated.

%% file: body.tex
\section{Introduction}
\label{sec:introduction}

\begin{figure*}[tbh]
    \centering
    \begin{subfigure}[tb]{0.25\linewidth}
        \centering
        \includegraphics[width=\linewidth]{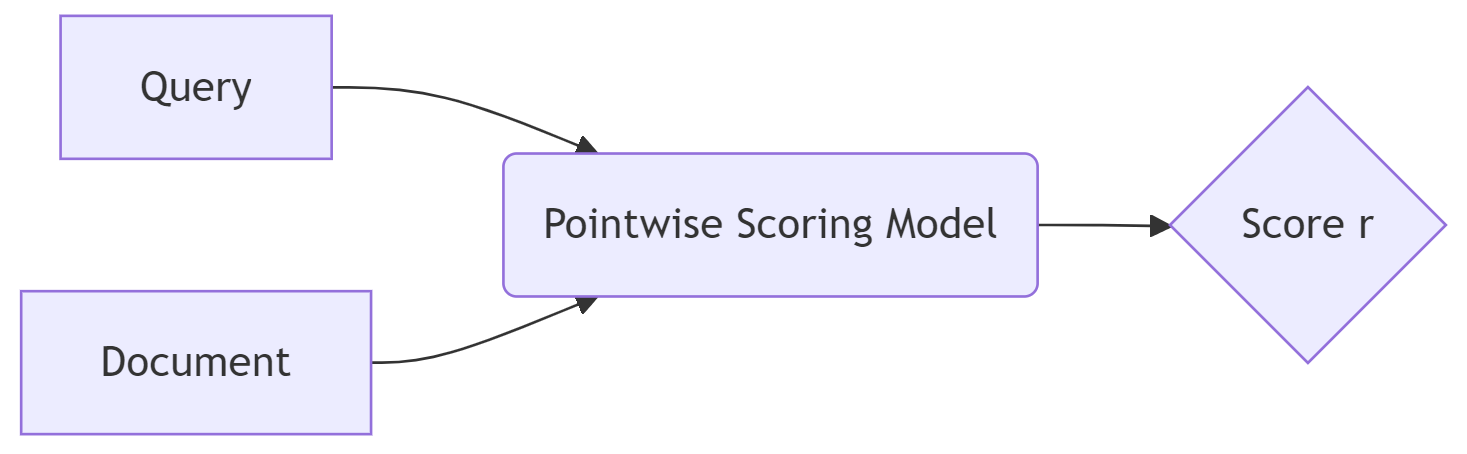}
        \caption{Pointwise Scoring}
        \label{fig:pointwise_scoring}
    \end{subfigure}
    \begin{subfigure}[tb]{0.35\linewidth}
        \centering
        \includegraphics[width=\linewidth]{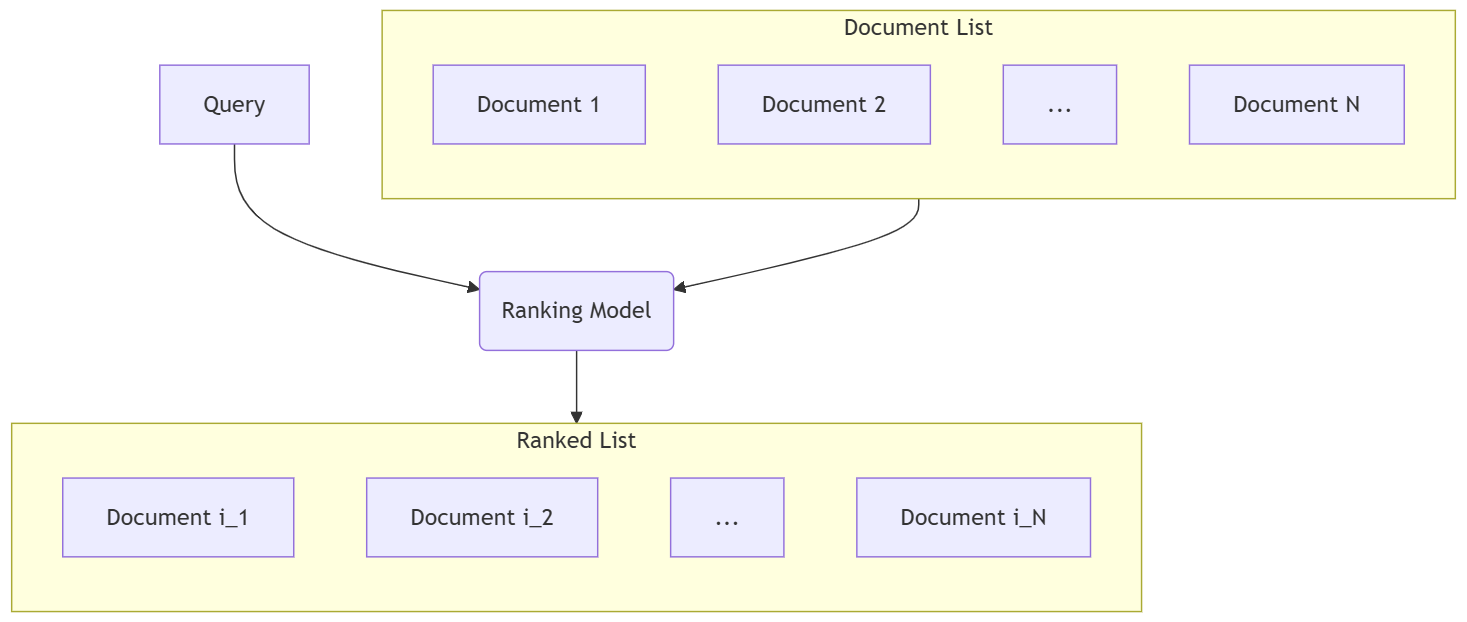}
        \caption{Ranking}
        \label{fig:ranking}
    \end{subfigure}
    \hfill
    \begin{subfigure}[tb]{0.35\linewidth}
        \centering
        \includegraphics[width=\linewidth]{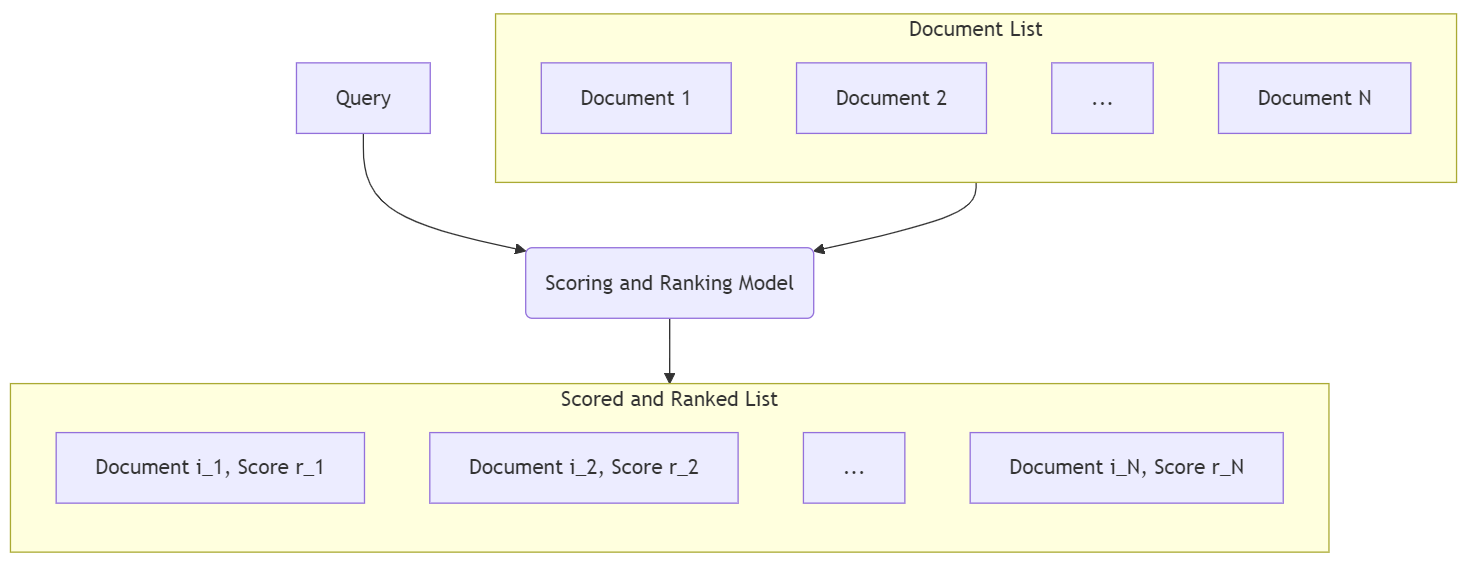}
        \caption{Scoring and Ranking}
        \label{fig:scoring_and_ranking}
    \end{subfigure}
    \caption{Comparison of Pointwise Scoring (left), Ranking (middle) and Scoring and Ranking (right). Here \(i_1, i_2, \dots, i_N\) is a permutation of \(1, 2, \dots, N\) and \(r_1, r_2, \dots, r_N\) are relevance scores.}
    \label{fig:comparison}
\end{figure*}

\begin{table*}[tb]
    \caption{Absolute NDCG@10 for pointwise scoring on ordinal relevance scales
        of varying size and listwise ranking (bubble), using GPT 4 Omni. See
        \cref{sec:experiments} for details.}
    \centering
    \begin{tabular}{lllllllllll}
        \toprule
        benchmark        & dl19  & dl20  & covid & nfc   & touche & dbpedia & scifact & signal & news  & robust04 \\
        method           &       &       &       &       &        &         &         &        &       &          \\
        \midrule
        2-point          & 0.645 & 0.636 & 0.776 & 0.371 & 0.427  & 0.395   & 0.706   & 0.333  & 0.429 & 0.547    \\
        3-point          & 0.711 & 0.685 & 0.814 & 0.387 & 0.417  & 0.427   & 0.730   & 0.338  & 0.472 & 0.597    \\
        5-point          & 0.736 & 0.698 & 0.820 & 0.390 & 0.400  & 0.437   & 0.744   & 0.343  & 0.479 & 0.630    \\
        7-point          & 0.747 & 0.709 & 0.851 & 0.399 & 0.387  & 0.436   & 0.758   & 0.337  & 0.484 & 0.634    \\
        11-point         & 0.737 & 0.697 & 0.849 & 0.395 & 0.359  & 0.440   & 0.753   & 0.351  & 0.487 & 0.638    \\
        ranking (bubble) & 0.747 & 0.700 & 0.852 & 0.404 & 0.373  & 0.470   & 0.799   & 0.334  & 0.531 & 0.639    \\
        \bottomrule
    \end{tabular}
    \label{fig:scale_ablation_ndcg10sweep}
\end{table*}

\begin{figure}[tb]
    \centering
    \includegraphics[width=\linewidth]{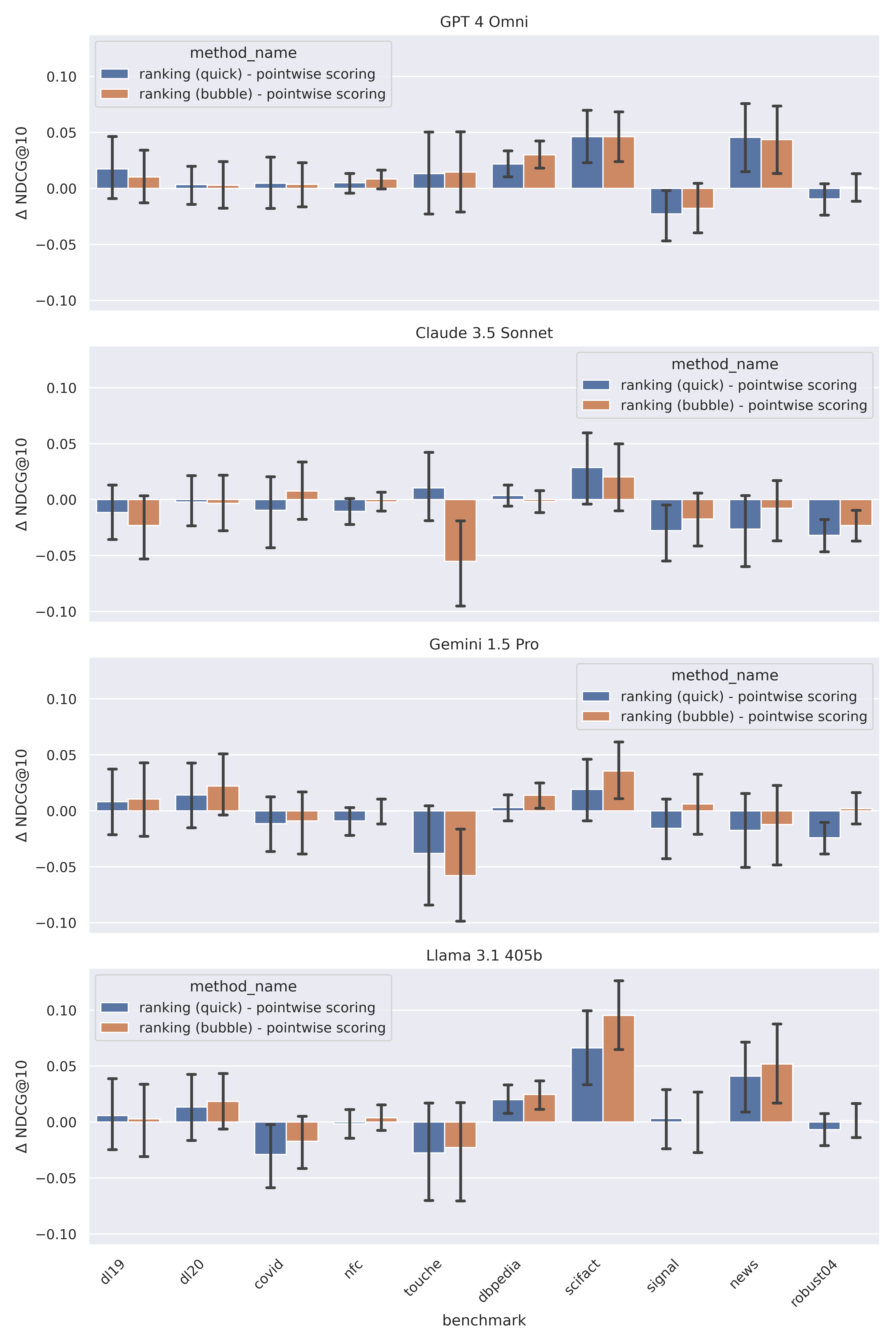}
    \caption{Observed differences in NDCG@10 for pointwise scoring and ranking
        on benchmark datasets. Bootstrap 95\% confidence intervals (see
        \cref{sec:quantifying-confidence}). For absolute values see see
        \cref{fig:ndcg10sweep}.}
    \label{fig:deltandcg10sweep}
\end{figure}

\begin{figure*}[tbh]
    \centering
    \begin{subfigure}[tb]{0.45\linewidth}
        \centering
        \includegraphics[width=\linewidth]{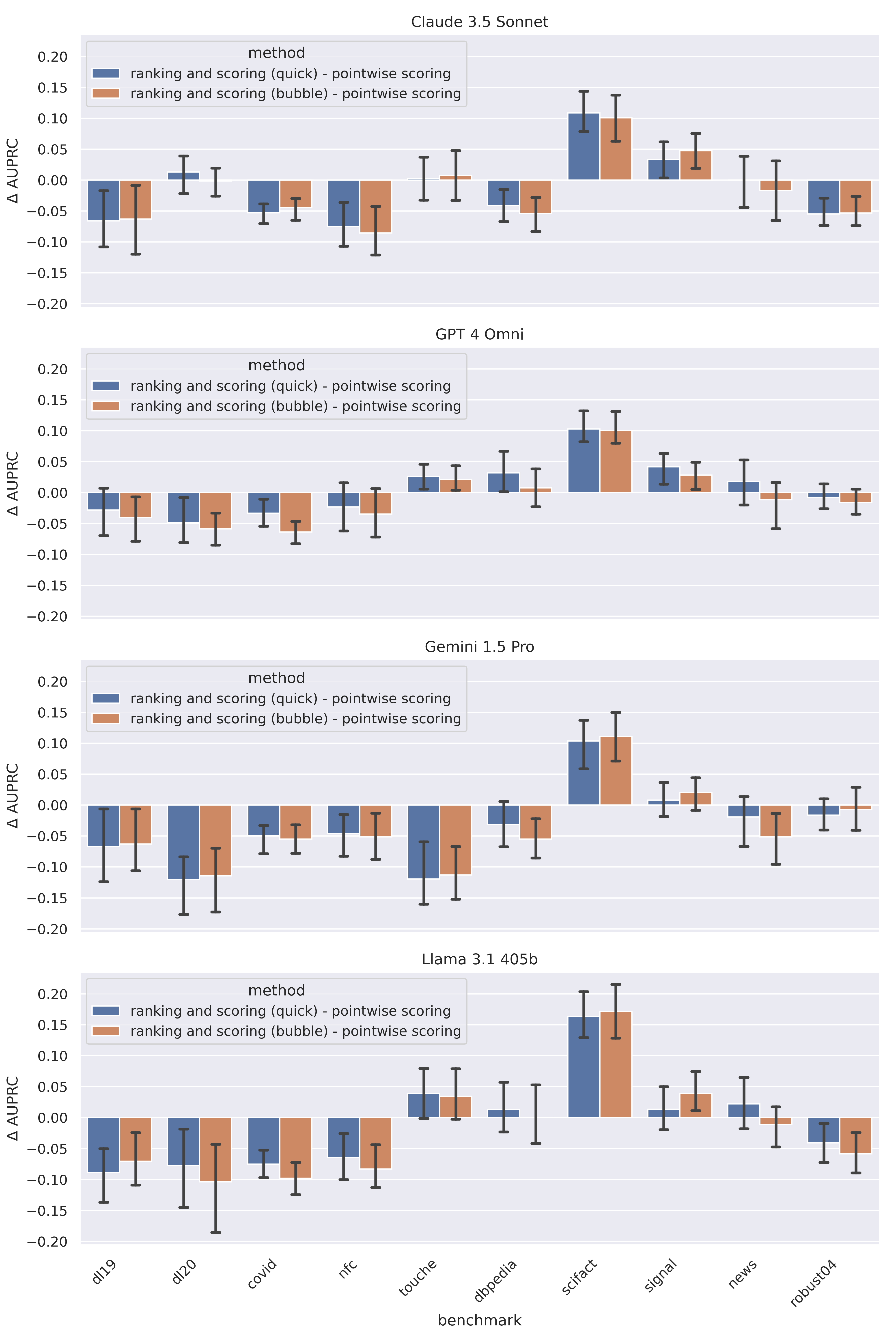}
        \caption{Observed differences in AUPRC for pointwise scoring and ranking on benchmark datasets. Bootstrap
            95\% confidence intervals (see \cref{sec:quantifying-confidence}).
            For absolute values see \cref{tab:auprcsweep}.}
        \label{fig:auprcsweep}
    \end{subfigure}
    \hfill
    \begin{subfigure}[tb]{0.45\linewidth}
        \centering
        \includegraphics[width=\linewidth]{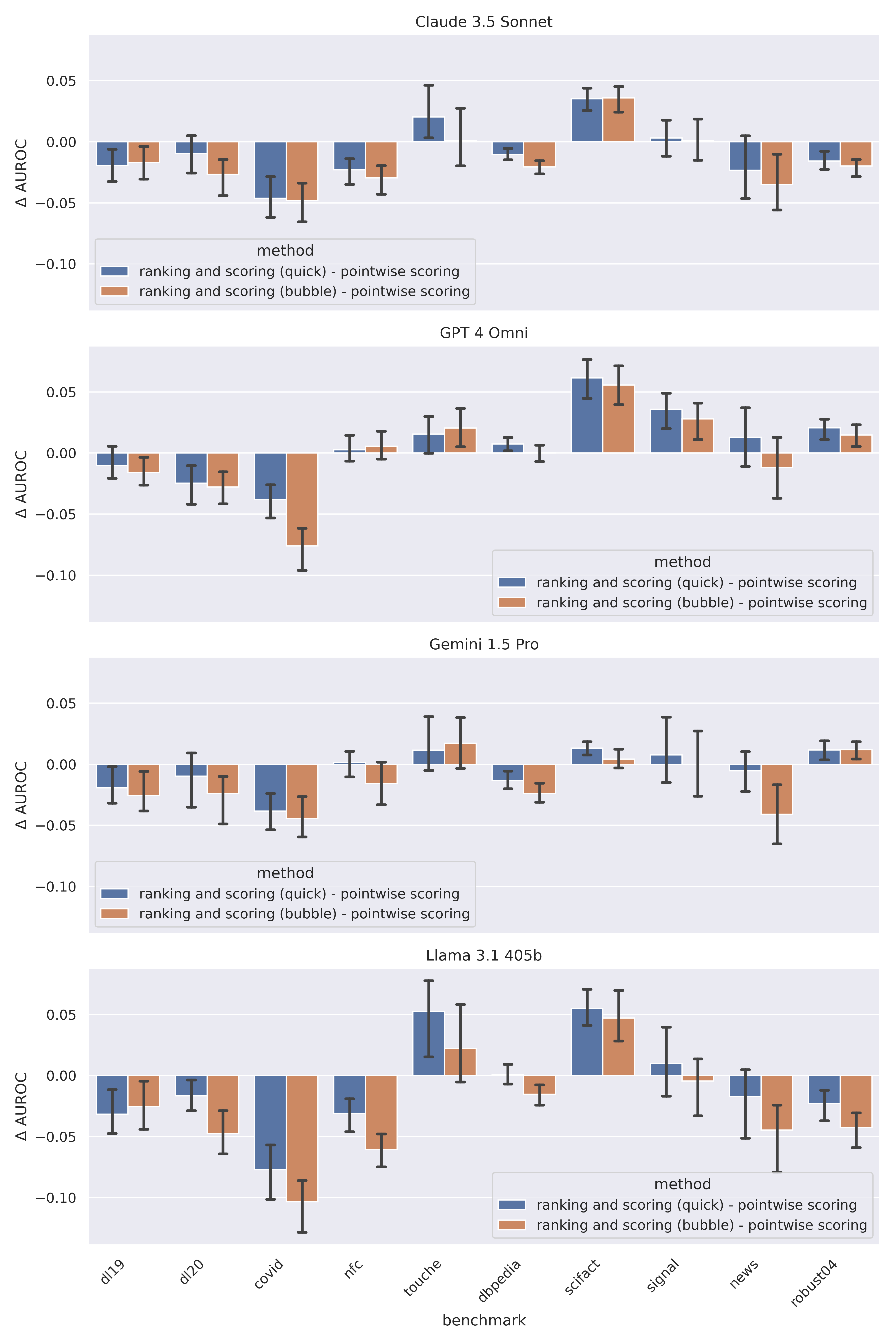}
        \caption{Observed differences in AUROC for pointwise scoring and ranking+scoring on benchmark datasets. Bootstrap
            95\% confidence intervals (see \cref{sec:quantifying-confidence}).
            For absolute values see \cref{tab:aurocsweep}.}
        \label{fig:aurocsweep}
    \end{subfigure}
    \caption{Comparing relevance classification performance of pointwise scoring vs. simultaneous ranking+scoring on benchmark datasets. Here we use an 11-point ordinal relevance scale.}
    \label{fig:classifiermetrics}
\end{figure*}

There is a long history in information retrieval of adding more powerful machine
learning models, as they become available, to the later phases of a search
pipeline.
Early examples of this pattern include learning to rank models based on
classical machine learning architectures like boosted trees. More recent
additions include cross-encoders fine-tuned from BERT or small (by today's
standards) GPT-style decoder-only transformers. The recent use of large language
models to rerank search results is a natural next step in this
progression.

If the above represents online inference-time use of LLMs in search, the offline
applications are just as promising. LLM-reranked search result lists serve as an
effectively limitless source of training data for learning to rank models, and
LLM relevance labels have proven effective for evaluating retrieval systems,
e.g. to select the best of several candidate text embedding models
\cite{khramtsovaLeveragingLLMsUnsupervised2024}. All of these applications, both
online and offline, depend on the quality of LLM relevance judgments.

Recently it was shown that prompting GPT-4 to rank lists of search results by
returning the permutation of indices that sorts the list in decreasing order of
relevance achieves state-of-the-art performance on information retrieval
benchmarks \cite{sunChatGPTGoodSearch2023}. Additionally, numerous experiments have found that
simply asking an LLM to rate whether or not a document is relevant to the query and
sorting on the basis of those binary relevance labels (or in some cases, on the
logit for the token ``Yes'') underperforms methods based on relative comparisons
of 2 or more documents and existing
non-LLM baselines \cite{sunChatGPTGoodSearch2023,maZeroShotListwiseDocument2023,qinLargeLanguageModels2023a,zhuangSetwiseApproachEffective2024}. This appears to have led to a prevailing view that LLMs are better at
making \emph{relative} rather than \emph{absolute} relevance judgments (see
\cref{sec:related-work}) -- we will refer to this as the \emph{LLM relative
    relevance judgments hypothesis}.
A fundamental bottleneck of using generative LLMs to sort lists of search
results is the limited context window, and hence limited number of search
results that can be sorted by any one LLM call. Fueled by the LLM relative
relevance judgments hypothesis a flurry of methods for
bootstrapping the ability to sort some fixed number of search results into
algorithms that sort longer lists has emerged. The basic theme is to pick any
classic sorting algorithm (bubble sort, quick sort, etc.) and adapt it
appropriately.

On the other hand, it has been found that the ranking performance obtained by
labeling individual search results by their relevance improves as one adds more
relevance labels. \citet{zhuangYesNoImproving2024} shows classifying
query-document pairs using a four bin relevance labeling scheme results in
significantly better rankings than a two bin relevance labeling scheme. A
small-scale ablation experiment in
\cite{guo2024generatingdiversecriteriaonthefly} shows performance may continue
to improve as one scales the number of relevance labels even further (with a
plateau observed beyond 11 or so labels)
\cite{guo2024generatingdiversecriteriaonthefly}. Neither of these papers include
a direct comparison with listwise ranking, but they make it obvious that prior
comparisons of listwise ranking with pointwise scoring underestimated the
capabilities of the latter.

We design experiments providing a direct and more fair comparison of LLM
pointwise scoring and listwise ranking, and find that simple pointwise scoring
prompts featuring relevance rubrics with up to 11 bins not only narrow the gap
with listwise ranking (see \cref{fig:scale_ablation_ndcg10sweep}), but do so to
a point where the difference in performance (according to commonly used metrics)
has low statistical significance for many model-dataset combinations.  In
evaluations including
\begin{itemize}
    \item 10 benchmark datasets from the BEIR and TREC
          Deep Learning suites
    \item the Claude 3.5 Sonnet, Gemini 1.5 Pro, GPT 4
          Omni, and Llama 3.1 405b models
    \item pointwise scoring with 11 bins and two representative listwise ranking methods,
          sliding window bubble sort and a multi-pivot early stopped quicksort, see \cref{sec:sortingalgorithms}
\end{itemize}
we find only 9 (of 40) model-dataset combination for which the pointwise
scoring method is significantly outperformed by a listwise ranking method in the
sense that the change in NDCG@10 metric is positive with 95\% confidence. The
situations where this outperformance occurs depend more heavily on the
underlying benchmark dataset (it is limitted to DBPedia, Scifact and News), as
opposed to the LLM or implementation details of the listwise ranking method.

In previous work on listwise ranking with LLMs, the model
output has been limited to a permutation, sorting the documents in decreasing
order of relevance.
Pointwise scoring methods have proven to be highly effective in important
applications distinct from ranking:
for example, when collecting synthetic relevance judgments to evaluate an
information retrieval system (see e.g.
\cite{khramtsovaLeveragingLLMsUnsupervised2024}), one requires relevance labels,
not just ranked lists. With the LLM relative relevance judgments hypothesis in
mind,\footnote{And looking at our information retrieval metrics showing listwise
    ranking methods continue to outperform pointwise scoring on some datasets.} it's
reasonable to ask whether relative comparisons between documents  can improve
LLM relevance \emph{labels}. We show that, via simple prompt modifications,
listwise ranking prompts can be upgraded to output both scores and ranks with no
significant impact on \emph{ranking} performance. We call these methods
\emph{ranking+scoring}.

Directly comparing relevance label quality, we find only 10/40 model-dataset
combinations where a listwise ranking increases the area under the
precision-recall curve (AUPRC) with 95\% confidence, and more strikingly 17/40
model-dataset combinations where both listwise ranking methods considered
\emph{decrease} AUPRC with 95\% confidence.
The area under the receiver operating characteristic curve (AUROC) metric tells a
similar story: a listwise ranking method results in a
significant improvement for 10/40 of the model-dataset combinations, and for
15/40 model-dataset combinations, both listwise ranking methods considered
significantly decrease AUROC.
We note that AUROC weights all
false positives equally, whereas AUPRC weights false positives at threshold
\(\tau\) by the inverse of the model's likelihood of outputting a relevance
score greater than \(\tau\) \cite{mcdermott2024closerlookaurocauprc} -- in the
terminology of information retrieval, this means AUPRC imposes a greater penalty
when the model places an irrelevant document towards the top of the search
results list.

To address data pollution concerns, we also create and evaluate on two novel
datasets with relevance labels collected after the training cut-off of the LLMs
evaluated. The underlying corpora are tax and accounting documents and passages
thereof, and the queries represent realistic (and in some cases real) questions
from professionals using an enterprise search system. We find that for all
model-benchmark combinations, listwise ranking significantly outperforms
pointwise scoring (again based on 95\% confidence of increase in NDCG@10, see \cref{fig:delta_ndcg10_proprietary}),
though for most models the advantage is slim. We leave an in-depth investigation
of what properties of a retrieval task (query distribution, corpus, etc.) make
LLM relative relevance judgments more valuable to future work.

We do not claim to refute a hypothesis that LLMs produce better relative
relevance judgments than absolute ones. However, our results demonstrate neither
standard information retrieval metrics like NDCG nor area-under-curve measures
of label quality on the standard benchmark data sets used in this research area
provide strong support for such a hypothesis in general (that is, without
qualifications, e.g. on the nature of the underlying distribution of queries and
corpus of documents).

\textbf{Main contributions}:
\begin{itemize}
    \item We demonstrate that prior work on LLM ranking has underestimated the
          performance of pointwise scoring relative to listwise ranking methods.
          Our comprehensive evaluation across 10 datasets and 4 models using
          minimal hyperparameter sweeps shows pointwise scoring with 11-bin
          relevance scales achieves competitive performance with listwise methods
          without extensive tuning.
    \item We present the first systematic evaluation of how relative comparisons
          impact LLM relevance \emph{labels} (rather than just rankings). Our
          ranking+scoring methods provide both permutations and per-document
          scores, enabling direct comparison of label quality. We find that
          listwise methods significantly decrease AUROC and AUPRC more often than
          they significantly increase these metrics.
    \item We emphasize the importance of comprehensive evaluation methodology,
          including confidence intervals for statistical significance testing,
          evaluation across multiple datasets and models, and consideration of
          both ranking performance and label quality metrics.
\end{itemize}

\section{Methods}

\subsection{Relevance judgment tasks}
\label{sec:implementations}

We consider three core relevance judgment tasks. The first, which we refer to as
\textit{pointwise scoring}, is to assign a query-document pair $(q,
    d)$ a relevance label $r \in \{0, 1, \ldots, R\}$ on a discrete ordinal
relevance scale. In our experiments the ordinal relevance scales are ranges of
non-negative integers (and in most of our experements $R = 10$).
The second, which we refer to as \emph{ranking}, is to take a query \(q\) and a list of
documents $(d_1, \ldots, d_N)$ and sort the list of documents in decreasing
order of relevance.  The third, which we refer to as \textit{ranking+scoring}, is to take a query $q$ and a list of documents $(d_1, \ldots, d_N)$
and simultaneously score each document $d_i$ with a relevance label $r_i \in \{0, 1, \ldots,
    R\}$ and sort the list of documents in decreasing order of relevance. See
\cref{fig:comparison} for diagrams of the three tasks.
Note that given relevance scores $r_1, \dots, r_N \in \{0, 1, \ldots, R\}$, for
all documents in a list $(d_1, \ldots, d_N)$ , whether obtained by pointwise
scoring or ranking+scoring, we can sort the list in decreasing order of
relevance. This is what we do in the case of pointwise scoring -- however, in the case
of ranking+scoring, the prevailing \emph{hypothesis} (see
\cref{sec:related-work}) is that using the sort order provides better relevance
sorting than using the scores alone.

In our experiments we implement pointwise scoring, ranking and ranking+scoring as zero-shot prompts for LLMs. See \cref{sec:prompttemplates} for (a
sample of) the prompt templates used.

\subsection{Sorting algorithms}
\label{sec:sortingalgorithms}

A fundamental bottleneck encountered when implementing LLM ranking is the number
of documents $n \ll N$ that can be input to a model on any given forward pass
due to context window limitations.
Prior work investigates a number of ways to bootstrap a model that can rank
lists of size $n$ into an algorithm for ranking lists of size $N$. There are
essentially as many different ways of bootstrapping as there are classical
sorting algorithms (which fundamentally bootstrap binary comparisons to sort
lists of arbitrary size) -- many approaches have been tried, see
\cref{sec:related-work}. We defer precise technical descriptions of the methods
we evaluate to \cref{sec:methodsdetails} -- here we focus on our \emph{choice}
of sorting methods to include.

Our choices are guided by the following (in some cases subjective) desiderata for
a sorting algorithm used to extend LLM listwise ranking to longer lists:
\begin{description}
    \item[D1] Must be compatible with general purpose ranking. That is, must not
          explicitly target a specific retrieval  metric (e.g. Precision@10) at
          the expense of others (e.g. high recall).
          % Note that this rules out \cite{parryTopPartitioningEfficientListWise2024}, which
          % essentially assumes we know the max number of docs we care about in advance.
    \item[D2] Should not require access to model internals or token
          probabilities -- we are interested in evaluating models behind APIs where
          this is not fully supported.
    \item[D3] Must have runtime complexity linear in the number of documents to be
          ranked (or admit an ``early stopping'' variant that does).
          % Think this rules out heap sort.
    \item[D4] Must not use more than a few tokens per query-document pair.
          % This rules out \cite{guo2024generatingdiversecriteriaonthefly} which has
          % an upfront cost of 100s-1000s of tokens per query.
    \item[D5] Must not require ensembling to achieve competitive results.
    \item[D6] Should admit parallelization such that multiple
          query-document pairs can be processed simultaneously.
\end{description}
Note that pointwise scoring with a zero shot prompt collecting only an integer
relevance label as output clearly satisfies D1-6.

Bubblesort (following
\citet{sunChatGPTGoodSearch2023,maZeroShotListwiseDocument2023}), which ranks
documents and sliding windows starting at the bottom of the list, satisfies
D1,\footnote{The size of sliding window overlap and number of passes does imply some
    prior about how many documents one care about, i.e. the ``K'' of ``Metric@K,''
    but this method is still in principle capable of improving
    ranking quality throughout the entire list} D2, D3 (with a cap on the number of
bubbling up passes), D4-5, but not D6. We choose to include bubblesort despite
failure of D6 since it's the most widely evaluated LLM listwise ranking method
(see \cref{sec:related-work}).

Since bubblesort does not satisfy D6, we prioritize that item in our choice of
another listwise ranking method. As noted in
\citet{qinLargeLanguageModels2023a,parryTopPartitioningEfficientListWise2024,zhuangSetwiseApproachEffective2024},
quicksort-type sorting methods have the natural property that many relevance
comparisons can be made in parallel, hence satisfies D6. In our literature
review we did not find an existing quicksort-type LLM listwise ranking method
satisfying all of D1-D5 (see discussion in
\cref{sec:related-work}), so we implement a ``multi-pivot quicksort with
telescoping'' algorithm -- \cref{alg:quicksort} (in the appendix) describes the algorithm in
pseudocode and the surrounding text provides additional details. Here we briefly
note that for our multi-pivot quicksort with telescoping algorithm, D1 is
satisfied\footnote{Similarly to the case of bubblesort, the number of pivots and
    number of passes through the list do imply some prior about how many documents
    one cares about.} by design, D2 and D4 are satisfied since we use a prompting strategy
equivalent to that of \citet{sunChatGPTGoodSearch2023}, D3 is satisfied by
limiting the number of passes through the list, and we demonstrate D5 with our
experimental results.

\emph{Computational Efficiency}: Beyond performance considerations, the
choice between pointwise and listwise methods involves important efficiency
trade-offs. Pointwise scoring is embarrassingly parallel, allowing all $N$
documents to be scored simultaneously, whereas listwise methods require serial
processing (bubble sort) or more limited parallelism (quicksort). With our
experimental parameters, listwise ranking methods consume approximately 3.4
times the input and output tokens of pointwise scoring due to overlapping
windows and telescoping procedures. However, quicksort-based listwise ranking
achieves parallelism degrees approaching that of pointwise scoring when
constrained by typical API rate limits. Detailed analysis is provided in
\cref{sec:efficiencyconsiderations}.

\subsection{Quantifying confidence}
\label{sec:quantifying-confidence}

A standard modelling assumption in information retrieval research is that queries are
independent and identically distributed (IID). This forms the basis of our
approach to estimating confidence intervals for our experimental results (here
we draw some inspiration from
\citet{oosterhuisReliableConfidenceIntervals2024}).\footnote{There are many
    other sources of randomness occurring in our experiments. For example, the fact
    that LLM outputs remain somewhat non-deterministic even though we run inference at zero
    temperature throughout. We do not attempt to quantify the impact of these other
    sources of randomness.}

To be precise, let \(D\) be an information retrieval dataset (e.g. one of the
benchmark datasets in our experiments) let \(Q\) be the set of queries occuring
in \(D\), and let \(N\) be the cardinality of \(Q\). Let \(m(q) \) be a metric
depending on the query \(q\). The primary example of such a metric considered in
this paper is
\begin{equation}
    \begin{split}
         & \text{NDCG@10}(\text{Ranking Method A}, q)    \\
         & -  \text{NDCG@10}(\text{Ranking Method B}, q)
    \end{split}
\end{equation}
i.e. the absolute difference in query-level NDCG@10 for two ranking methods A
and B.

To approximate the distribution of an estimate for \(E[m(q)]\) via the empirical
mean of \(N \) samples, we use the bootstrap method. For each query \(q\), we
sample \(N\) queries, say \(Q' \), with replacement from \(Q\) and
compute the empirical mean of \(m(q) \) over \(Q' \). We repeat this
process \(B\) times to form a bootstrap distribution of the empirical mean of
the metric. We may then use the bootstrap distribution to estimate statistics,
such as a 95\% confidence interval for  \(E[m(q)]\). \footnote{For most plots in this
    paper, this is accomplished by simply using defaults from the \texttt{seaborn}
    library in Python. We refer to the documentation on
    \href{https://seaborn.pydata.org/tutorial/error_bars.html}{Seaborn's default
        error bar method \texttt{'ci'}}.}

\section{Models}
\label{sec:models}

We implement \emph{pointwise scoring}, \emph{ranking} and \emph{ranking+scoring}
as prompts for the Anthropic Claude 3.5 Sonnet
\cite{anthropicClaude35Sonnet2024}, OpenAI GPT 4 Omni \cite{GPT4oSystemCarda},
Gemini 1.5 Pro \cite{teamGemini15Unlocking2024} and Llama 3.1 405b
\cite{grattafioriLlama3Herd2024} LLMs. In the case of GPT 4 Omni and Gemini 1.5
Pro, we use structured output generation to ensure adherance to a prescribed
JSON schema. For Claude 3.5 Sonnet and Llama 3.1 405b, structured outputs were
not available at the time of running experiments. For details on how we handle
errors including schema adherance failures, see
\cref{sec:handlingllminferenceerrors}.

The motivations behind our selection of models include:
\begin{description}
    \item[M1] Our focus is primarily on information retrieval performance and
          relevance classification accuracy, and our experiments show that these
          models obtain performance competitive with state of the art.
    \item[M2] Our interest is \emph{not primarily} in efficiency as measured by
          inference latency or cost per query, etc. Prior work has generally
          reported lower benchmark results while \emph{both} using smaller models
          \emph{and} adhering to our desiderata D1-D6.
    \item[M3] Prohibitive cost of evaluating on GPT 4 itself, Claude 3 Opus,
          Gemini 1.0 Ultra, and similar ruled out models one tier larger than
          those evaluated.
    \item[M4] We include several models to assess dataset vs. model dependence
          of performance trends.
\end{description}

\section{Datasets}
\label{sec:datasets}

For comparison with prior work, the following publicly available datasets are
included in our evaluations: TREC Deep Learning 2019 (TREC-DL19)
\cite{craswellOverviewTREC20192020}, TREC Deep Learning 2020 (TREC-DL20)
\cite{craswellOverviewTREC20202021}, and the subset of the Benchmarking-IR (BEIR)
collection \cite{thakurBEIRHeterogenousBenchmark2021} consisting of the Covid,
NFC, Touche, DBPedia, SciFact, Signal, News and Robust04 datasets, i.e. those
with less than 1000 queries.

\subsection{Novel benchmark datasets for retrieval of tax and accounting documents}
\label{sec:proprietarydatasets}

\begin{table}[tbh]
    \centering
    \caption{LLM models evaluated and their pre-training data cut-off dates}
    \label{tab:llm_training_data}
    \small{\begin{tabular}{ll}
            \toprule
            \textbf{LLM model} & \textbf{Pre-training data cut-off date} \\
            \midrule
            Claude 3.5 Sonnet  & April 2024                              \\
            GPT 4 Omni         & October, 2023                           \\
            Gemini 1.5 Pro     & December 2023 (estimate, see text)      \\
            Llama 3.1 405b     & December 2023                           \\
            \bottomrule
        \end{tabular}}
\end{table}

\begin{table*}[htb]
    \centering
    \caption{Summary statistics of datasets used in our experiments.}
    \small{\begin{tabular}{lrrrr}
            \toprule
            Benchmark                      & \# Docs                           & Median Doc Words & \#
            Queries                        & Median Relevance Judgements/Query                               \\
            \midrule
            Tax\&Accounting Full Documents & 4189                              & 461              & 334 & 20 \\
            Tax\&Accounting Passages       & 11534                             & 304              & 334 & 40 \\
            \bottomrule
        \end{tabular}}
    \label{tab:dataset_statistics}
\end{table*}

Endemic to evaluation of LLMs on publicly available benchmark
data is the concern that the models may have been trained on an appreciable fraction of the
entire Internet, and it is entirely possible that they have ``seen'' the
benchmark data (including its test split). This leaves open to question whether
the experiments above measure model capabilities generalizing to new data or
simply memorization, i.e. extraction of knowledge from parametric
memory.\footnote{Note that \cite{maZeroShotListwiseDocument2023} asserts data
    pollution concerns are orthogonal comparing the relative performance of
    pointwise scoring and listwise ranking; we do not agree. Consider the following
    thought experiment: take a novel ranking benchmark created entirely after the
    training cut-off of an LLM. Suppose that pointwise scoring and listwise
    ranking exhibit different performance on this benchmark.  Given the well
    established memorization capabilities of large language models, one would expect
    that after fine-tuning on the test split of the novel benchmark \emph{both}
    pointwise scoring and listwise ranking would achieve near perfect, and hence
    nearly identical accuracy. Hence it's possible for data pollution to make the
    performance of the two methods appear more similar than it would have been
    absent data pollution.}

There are multiple ways to ensure that evaluation data is novel: one is to use
data unlikely to have been readily accessible on the Internet prior to the time
of evaluation -- \cref{tab:llm_training_data} lists the pre-training data cut-offs
of LLMs used in our evaluations.\footnote{With the exception that for Gemini 1.5 Pro, our
    estimate is only an upper bound on the pre-training data cut-off of
    \texttt{gemini-1-5-pro-001}, deduced from the statement ``For our multilingual
    evaluations we use ... a machine translation benchmark [citation of
            \emph{Findings of the 2023 conference on machine translation (WMT23): LLMs are
                here but not quite there yet}, which appeared in December 2023] which was
    constructed after the model’s training data cut-off ...'' appearing in
    \cite{teamGemini15Unlocking2024}} Another is to use data that was \emph{created}
after the pre-training data cut-off of the LLMs. We employ both strategies to
construct two novel, proprietary information retrieval benchmark datasets. While
we are unfortunately unable to release the datasets themselves, in this section
we present a detailed description of the datasets results of our evaluations on
them.

Both datasets share the same corpus of documents and set of queries; the only
difference between the two is that, for one, we obtain relevance labels from
human experts at the full document level, whereas, for the other, we obtain
relevance labels at the passage (roughly 300 word chunk of a document) level.
That is, the full document and passage variants are designed to measure
different levels of retrieval granularity. The underlying corpus consists of
documents about federal tax law in the United States of America. The 334 queries
are assembled from a mix of real user queries entered into an enterprise
research system (both common queries and randomly sampled queries) and queries
written by domain experts. Annotation was completed during May 2024.
\Cref{sec:datasetconstruction} contains a detailed description of the dataset
creation process. In what follows, these datasets will be referred to as
\emph{Tax\&Accounting Full Document} and \emph{Tax\&Accounting Passage},
respectively. \Cref{tab:dataset_statistics} contains summary statistics of the
datasets.

\section{Experiments}
\label{sec:experiments}

\begin{figure}[tb]
    \centering
    \includegraphics[width=\linewidth]{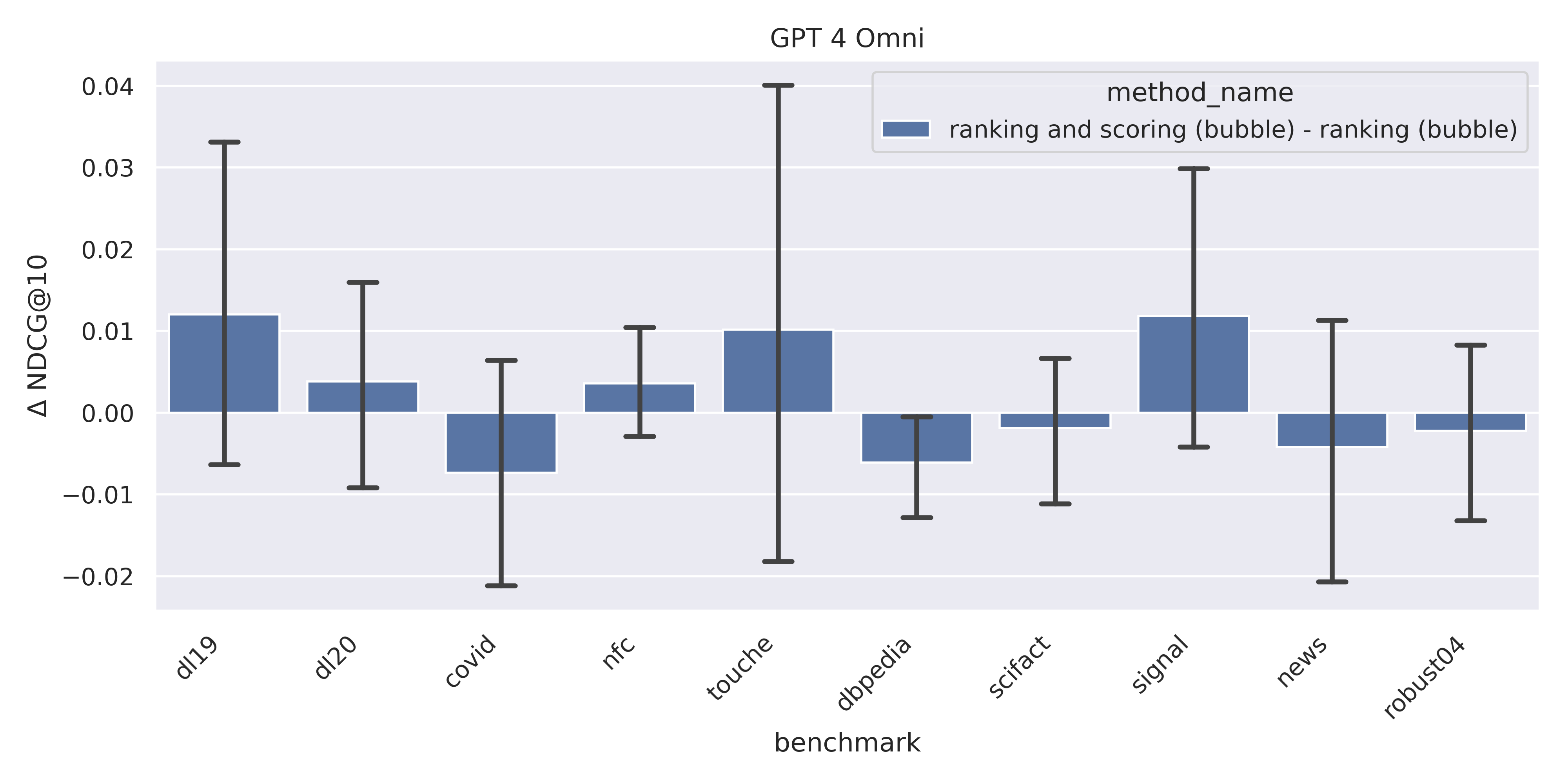}
    \caption{Observed differences in NDCG@10 for ranking+scoring vs. pure
        permutation generation listwise ranking (bubble-sort implementation) on
        benchmark datasets. Bootstrap
        95\% confidence intervals (see \cref{sec:quantifying-confidence}). For
        absolute NDCG@10 see \cref{fig:ndcg10sweep}.}
    \label{fig:bubble_ablation_ndcg10sweep}
\end{figure}

For each query, we collect the top 100 BM25 hits using Pyserini
\cite{Lin_etal_SIGIR2021_Pyserini}. For the publicly available benchmarks we
use Pyserini's prebuilt indices, and for the \emph{Tax\&Accounting} datasets we
build Pyserini indices following the official documentation. All documents and
passages are truncated to their first 300 words, following \cite{sunChatGPTGoodSearch2023}.

In the case of pointwise scoring, each query-document pair is labeled
independently. For the bubble sort and quicksort methods, we employ the
telescoping procedure described in \cref{sec:methodsdetails} with truncation at
50 documents followed by truncation at 20 documents. Unless otherwise specified,
an 11 point ordinal relevance scale (0-10) is used for all pointwise scoring and
ranking+scoring methods.

Following prior work, we compute NDCG metrics using the canonical
\href{https://github.com/usnistgov/trec_eval}{\texttt{trec\_eval}} tool, via the
convenient \href{https://github.com/cvangysel/pytrec_eval}{\texttt{pytrec\_eval}} interface.

\subsection{Results on public benchmarks}
\label{sec:publicbenchmarks}

In our first experiment, we reproduce an earlier finding that pointwise scoring
information retrieval improvemes with larger ordinal relevance scales
\cite{zhuangYesNoImproving2024,guo2024generatingdiversecriteriaonthefly}.  We
restrict attention to the GPT 4 Omni model and publicly available benchmark data
sets, and evaluate pointwise scoring using a variable number of relevance labels
(2, 3, 5, 7 or 11) and listwise ranking (bubble sort implementation).
\cref{fig:scale_ablation_ndcg10sweep} shows that, with the exception of the
Touche dataset, the performance of pointwise scoring as measured by NDCG@10
increases roughly monotonically with the size of the ordinal relevance scale; in
some cases a slight decrease in NDCG@10 is observed between 7 and 11 relevance
labels.\footnote{Touche is an obvious outlier in many of our experimental
    results; this possibly stems from a mismatch between our generic ``rank/score
    search results by relevance'' prompts applied uniformly across datasets and the
    more nuanced argument retrieval task the Touche dataset was constructed to
    evaluate.} Moreover, \cref{fig:scale_ablation_delta_ndcg10} shows that there are
only 3 (of 10) datasets considered, namely DBPedia, SciFact and News, for which
NDCG@10 performance of bubble sort listwise ranking  exceeds that of pointwise
scoring (using 11 relevance labels) with 95\% confidence. Hence in contrast with
prior work, we observe that pointwise scoring with 11 bins is \emph{competitive} with
listwise ranking across various datasets.

Next, we restrict attention to an 11-point ordinal relevance scale for pointwise
scoring and show the finding of \cref{fig:scale_ablation_ndcg10sweep} that
pointwise scoring with 11 bins is competitive with listwise ranking
qualitatively generalizes across LLMs and listwise ranking implementations. In
\cref{fig:deltandcg10sweep}, we only see listwise ranking significantly improve over
pointwise scoring as measured by positive change in NDCG@10 with 95\% confidence
for 9 out of 40 model-dataset combinations. In our experimental results this
condition is essentially insensitive to the question of whether listwise ranking is
implemented using bubble or quicksort (the lone exception being Gemini 1.5 Pro
on the SciFact dataset). Moreover, significant improvement in NDCG@10 is only observed
on the DBPedia, SciFact and News datasets. This suggests that the criteria for
listwise ranking methods to outperform pointwise scoring depends more heavily
on the underlying information retrieval data than the LLM or the implementation
details of the listwise ranking. For absolute values of NDCG@10 to accompany the
differences plotted in \cref{fig:deltandcg10sweep} see \cref{fig:ndcg10sweep}.

As mentioned in the introduction, prior work on listwise
ranking with LLMs has used pure permutation generation (ranking but not
scoring).
In \cref{fig:bubble_ablation_ndcg10sweep} we see that using GPT 4 Omni and
bubble-sort based methods, for 9/10 datasets considered, simultaneous
ranking+scoring does \emph{not} exhibit performance significantly different from
pure permutation generation ranking. There is a single dataset (DBPedia) where
simultaneous ranking+scoring decreases NDCG@10 with 95\% confidence, and in this
case the expected decrease in NDCG@10 is less than 0.01 (compare with the
absolute NDCG@10 values in \cref{fig:ndcg10sweep}). For the remaining datasets,
the difference in NDCG@10 is not significant.

Our experimental results discussed so far show that \emph{information retrieval
    performance} of listwise ranking significantly exceeds that of pointwise scoring
(with an 11-point scale) for less than a quarter of the model-dataset
combinations we consider, and that the performance of listwise ranking is not
significantly impacted if we upgrade pure permutation generation prompts to
return per document  relevance labels in addition to ranks. A natural follow
up question to ask is how the quality of relevance labels produced by
simultaneous ranking+scoring compared to those of pointwise scoring. In
other words, we ask: how do pointwise scoring and simultaneous ranking+scoring compare as
relevance classifiers?

\Cref{fig:auprcsweep} and \cref{fig:aurocsweep} show the differences in AUPRC
and AUROC, respectively, for pointwise scoring and ranking+scoring methods. To
compute these metrics, we first binarize ground truth relevance labels using the
documentation of the benchmark datasets. Specifically, for the MS MARCO passage
derived datasets (TREC DL19, TREC DL20) we declare a document relevant if its
ground truth label is at least 2, and for the other datasets we declare a
document relevant if its ground truth label is at least 1. At this point for
each query-document pair \(q, d \) we have a binary relevance label \(y \in \{0,
1\} \). We then divide the LLM relevance score \(r \) by 10 to obtain a score in
\([0,1]\), and use \texttt{scikit-learn}'s metrics library to compute AUPRC and
AUROC.

In \cref{fig:auprcsweep} we see 10 model-dataset combinations where a listwise
ranking method significantly increases AUPRC with 95\% confidence, and 15
combinations where both listwise ranking methods decrease AUPRC with 95\%
confidence; for the remaining 15 combinations the difference is not significant.
In \cref{fig:aurocsweep} we see 10 model-dataset combinations where a listwise
ranking method significantly increases AUROC with 95\% confidence, and 17
combinations where both listwise ranking methods decrease AUROC with 95\%
confidence; again for the remaining combinations the difference is not
significant. Note that trends in this plot are more dataset dependent than model
dependent -- for example, pointwise scoring outperforms listwise ranking as
measured by AUPRC/AUROC for all models on the Covid dataset.

\subsection{Results on Tax\&Accounting datasets}
\label{sec:proprietaryresults}

\begin{figure}[tb]
    \centering
    \includegraphics[width=\linewidth]{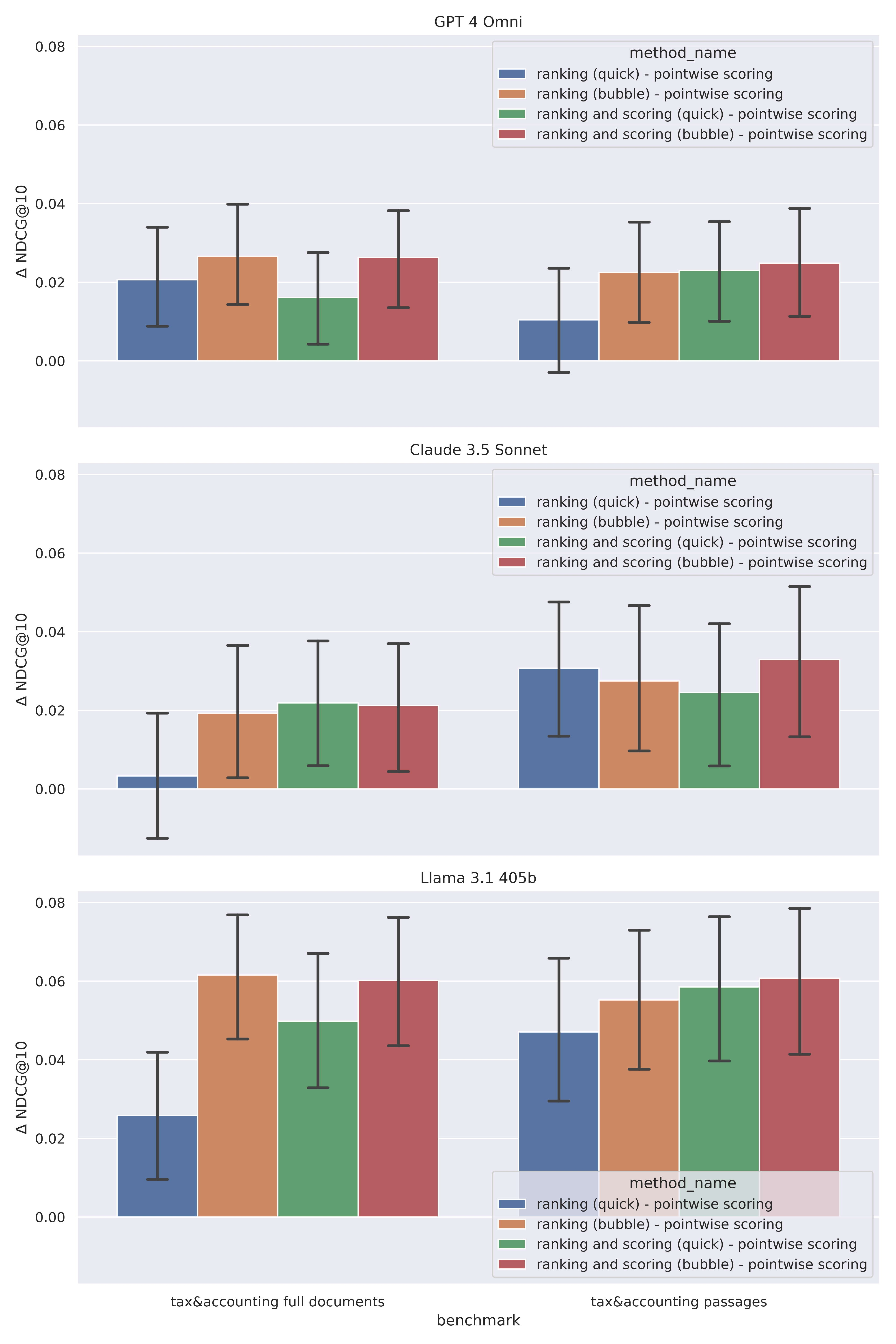}
    \caption{Observed differences in NDCG@10 for pointwise scoring and listwise
        ranking methods on Tax\&Accounting \{Full Document, Passage\} datasets. Bootstrap
        95\% confidence intervals (see \cref{sec:quantifying-confidence}). For
        absolute NDCG@10 see \cref{fig:ndcg10_proprietary}.}
    \label{fig:delta_ndcg10_proprietary}
\end{figure}

In \cref{fig:delta_ndcg10_proprietary} we see that, with the exception of the Full
Document dataset, Claude 3.5 Sonnet model and quicksort-based
ranking (without scoring), listwise ranking methods significantly outperform pointwise scoring
(using an 11-point ordinal relevance scale for all methods) with 95\% confidence.  Absolute values of NDCG@10 to accompany the differences plotted
in \cref{fig:delta_ndcg10_proprietary} are shown in \cref{fig:ndcg10_proprietary}, where we see
that for GPT 4 Omni and Claude 3.5 Sonnet the advantage of listwise ranking
methods is slim. Unfortunately
we were not able to obtain results for Gemini 1.5 Pro on the Tax\&Accounting
datasets by the time of writing. For AUPRC and AUROC results on the
Tax\&Accounting datasets we refer to \cref{sec:additionalresults}.

\section{Related work}
\label{sec:related-work-short}

See \cref{sec:related-work} for in-depth discussion; here we include a high-level summary.

\citet{sunChatGPTGoodSearch2023} and \citet{maZeroShotListwiseDocument2023}
independently proposed listwise ranking with LLMs, evaluating permutation
generation and finding it outperforms binary pointwise scoring.
\citet{sunChatGPTGoodSearch2023} further showed GPT-4 listwise ranking surpasses
state-of-the-art supervised models on numerous benchmarks.

Other notable listwise ranking approaches include
\citet{qinLargeLanguageModels2023a} studying pairwise ranking and sorting
algorithms like heapsort, and \citet{zhuangSetwiseApproachEffective2024}
proposing a ``setwise'' ranking cleverly leveraging token probabilities.

On the pointwise scoring side, \citet{zhuangYesNoImproving2024} demonstrated
that pointwise scoring using fine-grained relevance scales with more than two
labels significantly outperforms binary scoring. They considered scales of 2, 3
and 4 labels. \citet{guo2024generatingdiversecriteriaonthefly} introduced an
agentic approach called MCRanker, where the LLM generates multiple annotators
and relevance criteria, with the resulting scores then aggregated. They showed
performance improves with more relevance labels, plateauing after an 11-point
scale, however that experiment covered two datasets (Covid and Touche), their
prompt did not include a semantic relevance rubric, and resulted in lower
performance than that observed in our experimental results.

\citet{ni_diras_2025} explore pointwise vs. listwise ranking in the
context of Retrieval-Augmented Generation (RAG), proposing DIRAS to fine-tune
smaller LLMs for relevance annotation. Our work investigates the more foundational
question of pointwise vs. listwise ranking using zero-shot methods across
multiple datasets and models. See \cref{sec:related-work-extended} for a more detailed comparison.

%% file: appendix.tex
\section{Extended related work}
\label{sec:related-work}

To the best of our awareness \cite{sunChatGPTGoodSearch2023} and
\cite{maZeroShotListwiseDocument2023} independently proposed listwise ranking
with large language models.\footnote{First versions of papers appeared two weeks
    apart.} Both evaluate pure permutation generation listwise ranking (including a
bubble-sort extension to long lists of documents), finding that it outperforms
\emph{binary} pointwise scoring using  GPT-3 (\texttt{davinci-003}) on TREC-DL
benchmarks. The authors of \cite{sunChatGPTGoodSearch2023} go on to show that
listwise ranking performance of GPT-4 surpasses SOTA supervised models on
numerous benchmarks -- regarding the LLM relative relevance hypothesis, the
authors say
\begin{quote}
    First, from the result that [permutation generation] has significantly
    higher top-1 accuracy compared to other methods, we infer that LLMs can
    explicitly compare multiple passages with [permutation generation], allowing subtle differences
    between passages to be discerned. Second, LLMs gain a more comprehensive
    understanding of the query and passages by reading multiple passages with
    potentially complementary information, thus improving the model's ranking
    ability.
\end{quote}
and a similar hypothesis can be found in \cite{maZeroShotListwiseDocument2023}.

\citet{zhuangYesNoImproving2024} demonstrates that pointwise scoring using fine
grained relevance scales with more than two labels significantly outperforms
binary pointwise scoring. They consider ordinal relevance scales of size 2, 3
and 4 and evaluate FLAN PaLM2 \cite{anilPaLM2Technical2023} on the same BEIR
benchmarks used in this paper. One subtlety when comparing our experiments with
those of this paper is that those authors obtain better ranking performance when
computing a relevance score by \emph{averaging} over the probability
distribution of the ordinal relevance labels, whereas we simply take the most
likely label (in other words, they use a mean, we use a mode) -- in the language
of \cref{sec:implementations} this means that the best method of
\citet{zhuangYesNoImproving2024} does not satisfy desideratum D2.
This may explain why they observe NDCG@10 performance of pointwise scoring with
10 or more labels drops below that of binary scoring, whereas we observe roughly
monotonically increasing NDCG@10 with at most a slight drop going from 7 to 11
labels (but nowhere near a reversion to binary labelling performance). In
addition, the larger number of relevance labels we consider may mitigate what
those authors observe to be the largest shortcoming of simply using the mode,
namely that ``directly using the generated labels or scores results in lower ranking
performance compared to deriving scores from the log-likelihood, as it tends to
introduce ties between documents.'' Hence while our work is motivated by

\citet{qinLargeLanguageModels2023a} studies pairwise ranking, i.e. the limiting
case of listwise ranking where the number of documents is two. That work shows
that using only pairwise comparisons and open source language models of
reasonable size (FLAN-\{T5-XL, T5-XXL, UL2\}) can yield competitive and, for
some datasets and metrics, state-of-the-art performance. The authors consider
several methods for sorting large lists of documents, including ``Allpairs''
(essentially a round robin tournament comparing all pairs of distinct documents
in the list and then sorting by win ratio), heapsort and bubble sort with
sliding window of size two, overlap of size 1 and some fixed number of bubbling
up passes. Note that Allpairs and heapsort fail to satisfy desiderata D3 (their
runtime complexities are  respectively \(O(n^2)\) and \(O(n \log n)\)).
Moreover, while heapsort has better worst case complexity than quicksort and
shares the feature that tany comparisons can be executed in parallel, we discuss
in \cref{sec:methodsdetails} how ``early stopping'' multiple pivot quicksort
after a single pass through the list results in a coarse ranking, and we found
it (at least conceptually) difficult to see how similar behavior could be
achieved with heapsort. As such we could not see how to satisfy desiderata D3
using heapsort. The authors of \citet{qinLargeLanguageModels2023a} find
pairwise methods outperform pointwise scoring, though the pointwise scorers they
evaluate use only binary relevance labels.

\citet{parryTopPartitioningEfficientListWise2024} focuses on precision-oriented
top-k ranking and formulates it as a selection/partitioning problem. They
introduce a method called top down partitioning, where a single pivot document
is selected and then all remaining documents to be ranked are compared with the
pivot; documents deemed more relevant than the pivot are declared to be
``candidates'' and pointwise scored to obtain a final ranking. With evaluations
using GPT 3.5, LiT5 and Zephyr on TREC-DL \{19,20\}, Covid and Touche, they show
that this method is competitive with sliding window listwise ranking, reduces
the number of inference passes and increases the parallelism degree. One could
argue that our quicksort implementation is a zooming-in, multi-pivot version of
their top-down, single-pivot method. Note that the pre-specified budget of
candidates can be viewed as a hard prior on the number of documents to consider
(i.e. the ``K'' of a ``Metric@K'') encoded into top down partitioning. Moreover,
once the budget is exceeded, the method does not perform any ranking further
down the list. For this reason we do not consider top down partitioning to meet
desiderata D1. However, the idea of using a pivots and investigation into
sensitivity to choice of pivot in
\citet{parryTopPartitioningEfficientListWise2024} influenced our choices made
implementing multi-pivot quicksort.

\citet{zhuangSetwiseApproachEffective2024} introduces what they term ``setwise''
LLM list ranking, where a list of documents with unique identifiers is sent to
the model with a query, and the model is asked to return the identifier of the
most relevant document; the documents are then sorted according to the
probabilities of their unique identifiers. Note that not all LLM APIs used in
this paper permitted (at the time of running experiments) sufficient logit
access to test this method. The authors extend setwise ranking to long lists of
documents using bubble and heap sort algorithms (similar to our use of bubble
and quicksort). The evaluations use on Flan T5 \{L, XL, XXL\} models and the
TREC-DL \{19, 20\} and BEIR datasets evaluated in this paper, with a focus on
efficiency metrics.

\cite{guo2024generatingdiversecriteriaonthefly} introduces an agentic approach
to pointwise scoring called MCRanker, where for each query an LLM is prompted to
recruit multiple different human expert relevance annotators and then for each
annotator an LLM is prompted to come up with relevance criteria (criteria of a
passage that would make it relevant along with weights). Then each passage is
scored (on a scale of 0-10, as in our main experiments) by each annotator
according to their respective criteria and the resulting relevance scores are
aggregated. The authors evaluate GPT-4-1106-Preview on the same 8 BEIR datasets
appearing in our experiments. Note that even though the up-front cost of
generating annotator personalities and relevance criteria is amortized across
grading all passages, the total number of output tokens per query passage pair
generated by this pointwise scoring method is higher than the pointwise
scoring methods appearing in our experiments, and moreover, the annotator
personalities and criteria must be generated in serial. Hence we do not consider
this method to satisfy D4 and D6. Their paper includes an ablation on the Covid
and Touche data sets, demonstrating that their MCRanker performance improves as
the number of relevance labels increases, with a plateau after an 11-point
scale. They also include what they term ``direct'' pointwise scoring where the
LLM is prompted to return the relevance score of a passage on a 0-10 scale as a
single integer, but without any rubric describing what relevance level  each
integer represents. The performance we obtain with single integer output point
wise scores using a fixed (across all data sets and queries) semantic relevance
rubric is better than what they report for their ``direct''  method, and
comparable to the best of their MCRanker methods (although the fact that we use
GPT 4 Omni whereas they use GPT-4-1106-Preview makes direct comparison
difficult).

\citet{ni_diras_2025} explore pointwise vs. listwise ranking in the
context of Retrieval-Augmented Generation (RAG), proposing DIRAS to fine-tune
smaller LLMs for relevance annotation. While DIRAS focuses on RAG applications
and uses chain-of-thought prompting with GPT-4-generated per-query relevance
definitions, our work investigates the more foundational question of pointwise
vs. listwise ranking in information retrieval using zero-shot methods with
static prompts across datasets. DIRAS evaluates primarily on one custom dataset
and one model (GPT-4), whereas we conduct comprehensive evaluation across 10
datasets and multiple models with confidence intervals. Their binary relevance
approach combined with floating-point confidence scores may be similar in
effect to using large ordinal relevance scales, though it lacks the semantic
clarity of explicit relevance rubrics. Moreover, DIRAS uses chain-of-thought
prompting for both queries and query-document pairs, which is computationally
expensive compared to our methods, and does not report confidence intervals for
their results.

Finally, we note a number of other approaches to listwise ranking with LLMs
that, while not as directly comparable to our work, are nonetheless worth
mentioning. \citet{chenTourRankUtilizingLarge2024} formulates document ranking
as a relevance ``tournament,'' where documents are ranked in batches, and the
top documents within a batch are sent on to the next round, and a document gets
a point every time it progresses to another round (such points can be averaged
over multiple tournaments with different randomness injected). We omit this from
our evaluation since in their experiments it only surpassed e.g.
\cite{sunChatGPTGoodSearch2023} when provided over double the number of model
forward passes. \citet{nouriinanlooReRankingStepStep2024} evaluates a two-stage
pipeline where pointwise scoring is used as a relevance filter before a final
listwise ranking step, and shows this setup can dramatically boost the
performance of a small open source LLM, namely Mixtral-8x7B-Instruct (though not
to the point of being competitive with GPT-4). While we focus more on comparing
pointwise scoring and listwise ranking, this work is interesting in that it
shows that the two methods can be complementary.

\section{Additional experimental results}
\label{sec:additionalresults}

\begin{figure}[tb]
    \centering
    \includegraphics[width=\linewidth]{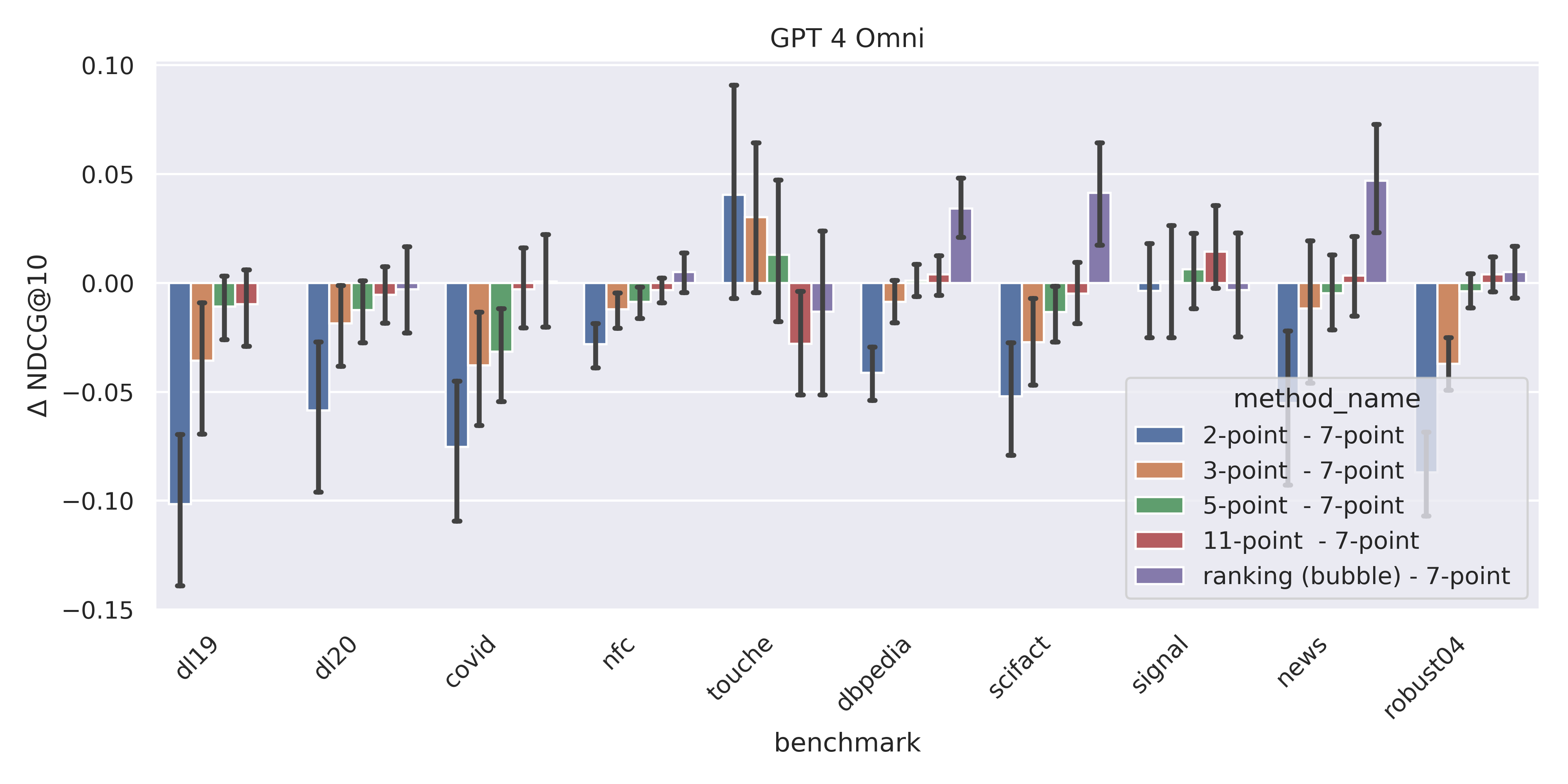}
    \caption{Change in NDCG@10 for pointwise scoring on ordinal relevance scales
        of varying size and listwise ranking (bubble), using GPT 4 Omni. Bootstrap
        95\% confidence intervals (see \cref{sec:quantifying-confidence}).}
    \label{fig:scale_ablation_delta_ndcg10}
\end{figure}

\begin{table}
    \centering
    \caption{Absolute NDCG@10 for pointwise scoring and listwise ranking on benchmark datasets.}
    \footnotesize{\begin{tabular}{llllllllllll}
            \toprule
                                                  & benchmark                & dl19  & dl20  & covid & nfc   & touche & dbpedia & scifact & signal & news  & robust04 \\
            model                                 & method                   &       &       &       &       &        &         &         &        &       &          \\
            \midrule
            \multirow[t]{5}{*}{GPT 4 Omni}        & pointwise scoring        & 0.737 & 0.697 & 0.849 & 0.395 & 0.359  & 0.440   & 0.753   & 0.351  & 0.487 & 0.638    \\
                                                  & ranking (bubble)         & 0.747 & 0.700 & 0.852 & 0.404 & 0.373  & 0.470   & 0.799   & 0.334  & 0.531 & 0.639    \\
                                                  & ranking (quick)          & 0.754 & 0.701 & 0.853 & 0.401 & 0.372  & 0.462   & 0.799   & 0.329  & 0.533 & 0.628    \\
                                                  & ranking+scoring (bubble) & 0.759 & 0.704 & 0.845 & 0.398 & 0.384  & 0.464   & 0.797   & 0.345  & 0.527 & 0.636    \\
                                                  & ranking+scoring (quick)  & 0.761 & 0.695 & 0.862 & 0.404 & 0.353  & 0.461   & 0.803   & 0.326  & 0.535 & 0.637    \\
            \cline{1-12}
            \multirow[t]{5}{*}{Claude 3.5 Sonnet} & pointwise scoring        & 0.746 & 0.700 & 0.849 & 0.396 & 0.385  & 0.469   & 0.763   & 0.351  & 0.543 & 0.621    \\
                                                  & ranking (bubble)         & 0.723 & 0.697 & 0.857 & 0.394 & 0.330  & 0.467   & 0.784   & 0.334  & 0.535 & 0.598    \\
                                                  & ranking (quick)          & 0.734 & 0.698 & 0.839 & 0.386 & 0.396  & 0.473   & 0.792   & 0.324  & 0.517 & 0.589    \\
                                                  & ranking+scoring (bubble) & 0.730 & 0.693 & 0.854 & 0.383 & 0.358  & 0.475   & 0.785   & 0.338  & 0.533 & 0.599    \\
                                                  & ranking+scoring (quick)  & 0.734 & 0.704 & 0.846 & 0.389 & 0.331  & 0.470   & 0.789   & 0.334  & 0.544 & 0.601    \\
            \cline{1-12}
            \multirow[t]{5}{*}{Gemini 1.5 Pro}    & pointwise scoring        & 0.720 & 0.658 & 0.845 & 0.391 & 0.385  & 0.441   & 0.758   & 0.321  & 0.488 & 0.616    \\
                                                  & ranking (bubble)         & 0.730 & 0.681 & 0.836 & 0.390 & 0.327  & 0.455   & 0.794   & 0.327  & 0.476 & 0.618    \\
                                                  & ranking (quick)          & 0.728 & 0.673 & 0.834 & 0.382 & 0.347  & 0.444   & 0.777   & 0.305  & 0.471 & 0.592    \\
                                                  & ranking+scoring (bubble) & 0.749 & 0.684 & 0.845 & 0.384 & 0.331  & 0.451   & 0.786   & 0.341  & 0.483 & 0.602    \\
                                                  & ranking+scoring (quick)  & 0.733 & 0.682 & 0.824 & 0.383 & 0.312  & 0.444   & 0.787   & 0.326  & 0.485 & 0.598    \\
            \cline{1-12}
            \multirow[t]{5}{*}{Llama 3.1 405b}    & pointwise scoring        & 0.737 & 0.692 & 0.858 & 0.396 & 0.414  & 0.437   & 0.691   & 0.311  & 0.466 & 0.608    \\
                                                  & ranking (bubble)         & 0.740 & 0.754 & 0.841 & 0.400 & 0.391  & 0.461   & 0.786   & 0.310  & 0.518 & 0.616    \\
                                                  & ranking (quick)          & 0.747 & 0.712 & 0.829 & 0.394 & 0.386  & 0.457   & 0.757   & 0.314  & 0.507 & 0.599    \\
                                                  & ranking+scoring (bubble) & 0.744 & 0.698 & 0.843 & 0.382 & 0.400  & 0.471   & 0.780   & 0.338  & 0.509 & 0.610    \\
                                                  & ranking+scoring (quick)  & 0.748 & 0.725 & 0.857 & 0.396 & 0.383  & 0.462   & 0.768   & 0.307  & 0.512 & 0.608    \\
            \cline{1-12}
            \bottomrule
        \end{tabular}}
    \label{fig:ndcg10sweep}
\end{table}

\begin{table}
    \centering
    \caption{Absolute NDCG@10 for pointwise scoring and listwise ranking on benchmark and Tax\&Accounting datasets.}
    \small{\begin{tabular}{llll}
            \toprule
                                                  & benchmark                    & tax\&accounting full documents & tax\&accounting passages \\
            model                                 & method                       &                                &                          \\
            \midrule
            \multirow[t]{5}{*}{GPT 4 Omni}        & pointwise scoring            & 0.619                          & 0.519                    \\
                                                  & ranking (bubble)             & 0.645                          & 0.541                    \\
                                                  & ranking (quick)              & 0.639                          & 0.529                    \\
                                                  & ranking and scoring (bubble) & 0.645                          & 0.544                    \\
                                                  & ranking and scoring (quick)  & 0.635                          & 0.542                    \\
            \cline{1-4}
            \multirow[t]{5}{*}{Claude 3.5 Sonnet} & pointwise scoring            & 0.600                          & 0.511                    \\
                                                  & ranking (bubble)             & 0.619                          & 0.538                    \\
                                                  & ranking (quick)              & 0.603                          & 0.542                    \\
                                                  & ranking and scoring (bubble) & 0.621                          & 0.544                    \\
                                                  & ranking and scoring (quick)  & 0.622                          & 0.535                    \\
            \cline{1-4}
            \multirow[t]{5}{*}{Llama 3.1 405b}    & pointwise scoring            & 0.564                          & 0.479                    \\
                                                  & ranking (bubble)             & 0.626                          & 0.534                    \\
                                                  & ranking (quick)              & 0.613                          & 0.526                    \\
                                                  & ranking and scoring (bubble) & 0.624                          & 0.540                    \\
                                                  & ranking and scoring (quick)  & 0.614                          & 0.537                    \\
            \cline{1-4}
            \bottomrule
        \end{tabular}}
    \label{fig:ndcg10_proprietary}
\end{table}

\begin{table}
    \centering
    \caption{Absolute AUPRC for pointwise scoring and ranking+scoring on benchmark datasets.}
    \footnotesize{\begin{tabular}{llllllllllll}
            \toprule
                                          & benchmark                    & covid & dbpedia & dl19  & dl20  & news  & nfc   & robust04 & scifact & signal & touche \\
            model                         & method                       &       &         &       &       &       &       &          &         &        &        \\
            \midrule
            \multirow[t]{3}{*}{anthropic} & pointwise scoring            & 0.884 & 0.565   & 0.765 & 0.162 & 0.552 & 0.499 & 0.606    & 0.537   & 0.315  & 0.209  \\
                                          & ranking and scoring (bubble) & 0.839 & 0.511   & 0.702 & 0.161 & 0.536 & 0.414 & 0.552    & 0.637   & 0.363  & 0.216  \\
                                          & ranking and scoring (quick)  & 0.831 & 0.524   & 0.699 & 0.175 & 0.552 & 0.424 & 0.551    & 0.645   & 0.348  & 0.212  \\
            \cline{1-12}
            \multirow[t]{3}{*}{llama}     & pointwise scoring            & 0.872 & 0.483   & 0.759 & 0.276 & 0.482 & 0.449 & 0.586    & 0.497   & 0.294  & 0.220  \\
                                          & ranking and scoring (bubble) & 0.774 & 0.483   & 0.688 & 0.172 & 0.470 & 0.366 & 0.527    & 0.669   & 0.333  & 0.255  \\
                                          & ranking and scoring (quick)  & 0.797 & 0.496   & 0.670 & 0.198 & 0.504 & 0.385 & 0.545    & 0.661   & 0.307  & 0.259  \\
            \cline{1-12}
            \multirow[t]{3}{*}{openai}    & pointwise scoring            & 0.876 & 0.486   & 0.789 & 0.261 & 0.516 & 0.467 & 0.605    & 0.551   & 0.303  & 0.194  \\
                                          & ranking and scoring (bubble) & 0.812 & 0.493   & 0.749 & 0.203 & 0.504 & 0.432 & 0.589    & 0.652   & 0.331  & 0.216  \\
                                          & ranking and scoring (quick)  & 0.843 & 0.518   & 0.761 & 0.212 & 0.534 & 0.444 & 0.597    & 0.653   & 0.345  & 0.220  \\
            \cline{1-12}
            \multirow[t]{3}{*}{vertex}    & pointwise scoring            & 0.869 & 0.522   & 0.762 & 0.284 & 0.530 & 0.465 & 0.572    & 0.552   & 0.327  & 0.295  \\
                                          & ranking and scoring (bubble) & 0.814 & 0.468   & 0.699 & 0.170 & 0.478 & 0.414 & 0.565    & 0.663   & 0.348  & 0.182  \\
                                          & ranking and scoring (quick)  & 0.820 & 0.491   & 0.695 & 0.164 & 0.511 & 0.419 & 0.555    & 0.656   & 0.335  & 0.175  \\
            \cline{1-12}
            \bottomrule
        \end{tabular}}
    \label{tab:auprcsweep}
\end{table}

\begin{table}
    \centering
    \caption{Absolute AUROC for pointwise scoring and ranking+scoring on benchmark datasets.}
    \footnotesize{\begin{tabular}{llllllllllll}
            \toprule
                                          & benchmark                    & covid & dbpedia & dl19  & dl20  & news  & nfc   & robust04 & scifact & signal & touche \\
            model                         & method                       &       &         &       &       &       &       &          &         &        &        \\
            \midrule
            \multirow[t]{3}{*}{anthropic} & pointwise scoring            & 0.893 & 0.903   & 0.932 & 0.878 & 0.859 & 0.864 & 0.850    & 0.951   & 0.805  & 0.676  \\
                                          & ranking and scoring (bubble) & 0.845 & 0.882   & 0.915 & 0.852 & 0.824 & 0.835 & 0.830    & 0.987   & 0.805  & 0.677  \\
                                          & ranking and scoring (quick)  & 0.847 & 0.892   & 0.913 & 0.868 & 0.836 & 0.842 & 0.834    & 0.986   & 0.808  & 0.696  \\
            \cline{1-12}
            \multirow[t]{3}{*}{llama}     & pointwise scoring            & 0.884 & 0.869   & 0.921 & 0.885 & 0.809 & 0.848 & 0.837    & 0.924   & 0.772  & 0.665  \\
                                          & ranking and scoring (bubble) & 0.780 & 0.854   & 0.895 & 0.838 & 0.764 & 0.787 & 0.795    & 0.971   & 0.768  & 0.687  \\
                                          & ranking and scoring (quick)  & 0.807 & 0.870   & 0.889 & 0.869 & 0.792 & 0.817 & 0.814    & 0.979   & 0.782  & 0.717  \\
            \cline{1-12}
            \multirow[t]{3}{*}{openai}    & pointwise scoring            & 0.888 & 0.875   & 0.935 & 0.892 & 0.813 & 0.838 & 0.824    & 0.924   & 0.768  & 0.673  \\
                                          & ranking and scoring (bubble) & 0.812 & 0.875   & 0.919 & 0.864 & 0.801 & 0.844 & 0.839    & 0.980   & 0.796  & 0.693  \\
                                          & ranking and scoring (quick)  & 0.850 & 0.882   & 0.925 & 0.867 & 0.826 & 0.841 & 0.845    & 0.986   & 0.804  & 0.688  \\
            \cline{1-12}
            \multirow[t]{3}{*}{vertex}    & pointwise scoring            & 0.878 & 0.882   & 0.922 & 0.871 & 0.829 & 0.842 & 0.828    & 0.970   & 0.775  & 0.657  \\
                                          & ranking and scoring (bubble) & 0.833 & 0.858   & 0.897 & 0.848 & 0.788 & 0.826 & 0.840    & 0.974   & 0.774  & 0.675  \\
                                          & ranking and scoring (quick)  & 0.840 & 0.869   & 0.903 & 0.862 & 0.823 & 0.843 & 0.840    & 0.983   & 0.782  & 0.669  \\
            \cline{1-12}
            \bottomrule
        \end{tabular}}
    \label{tab:aurocsweep}
\end{table}

\begin{table}
    \centering
    \caption{Absolute AUROC for pointwise scoring and ranking+scoring on Tax\&Accounting datasets.}
    \small{\begin{tabular}{llll}
            \toprule
                                          & benchmark                    & tax\&accounting full documents & tax\&accounting passages \\
            model                         & method                       &                                &                          \\
            \midrule
            \multirow[t]{3}{*}{anthropic} & pointwise scoring            & 0.893                          & 0.850                    \\
                                          & ranking and scoring (bubble) & 0.893                          & 0.860                    \\
                                          & ranking and scoring (quick)  & 0.898                          & 0.868                    \\
            \cline{1-4}
            \multirow[t]{3}{*}{llama}     & pointwise scoring            & 0.895                          & 0.836                    \\
                                          & ranking and scoring (bubble) & 0.880                          & 0.820                    \\
                                          & ranking and scoring (quick)  & 0.902                          & 0.844                    \\
            \cline{1-4}
            \multirow[t]{3}{*}{openai}    & pointwise scoring            & 0.892                          & 0.849                    \\
                                          & ranking and scoring (bubble) & 0.903                          & 0.852                    \\
                                          & ranking and scoring (quick)  & 0.910                          & 0.867                    \\
            \cline{1-4}
            \bottomrule
        \end{tabular}}
\end{table}

\begin{table}
    \centering
    \caption{Difference in AUPRC for pointwise scoring and listwise ranking on Tax\&Accounting datasets.}
    \begin{tabular}{llll}
        \toprule
                                      & benchmark                    & tax\&accounting full documents & tax\&accounting passages \\
        model                         & method                       &                                &                          \\
        \midrule
        \multirow[t]{3}{*}{anthropic} & pointwise scoring            & 0.893                          & 0.850                    \\
                                      & ranking and scoring (bubble) & 0.893                          & 0.860                    \\
                                      & ranking and scoring (quick)  & 0.898                          & 0.868                    \\
        \cline{1-4}
        \multirow[t]{3}{*}{llama}     & pointwise scoring            & 0.895                          & 0.836                    \\
                                      & ranking and scoring (bubble) & 0.880                          & 0.820                    \\
                                      & ranking and scoring (quick)  & 0.902                          & 0.844                    \\
        \cline{1-4}
        \multirow[t]{3}{*}{openai}    & pointwise scoring            & 0.892                          & 0.849                    \\
                                      & ranking and scoring (bubble) & 0.903                          & 0.852                    \\
                                      & ranking and scoring (quick)  & 0.910                          & 0.867                    \\
        \cline{1-4}
        \bottomrule
    \end{tabular}
\end{table}

\begin{figure}
    \centering
    \begin{subfigure}[tb]{0.45\linewidth}
        \includegraphics[width=\linewidth]{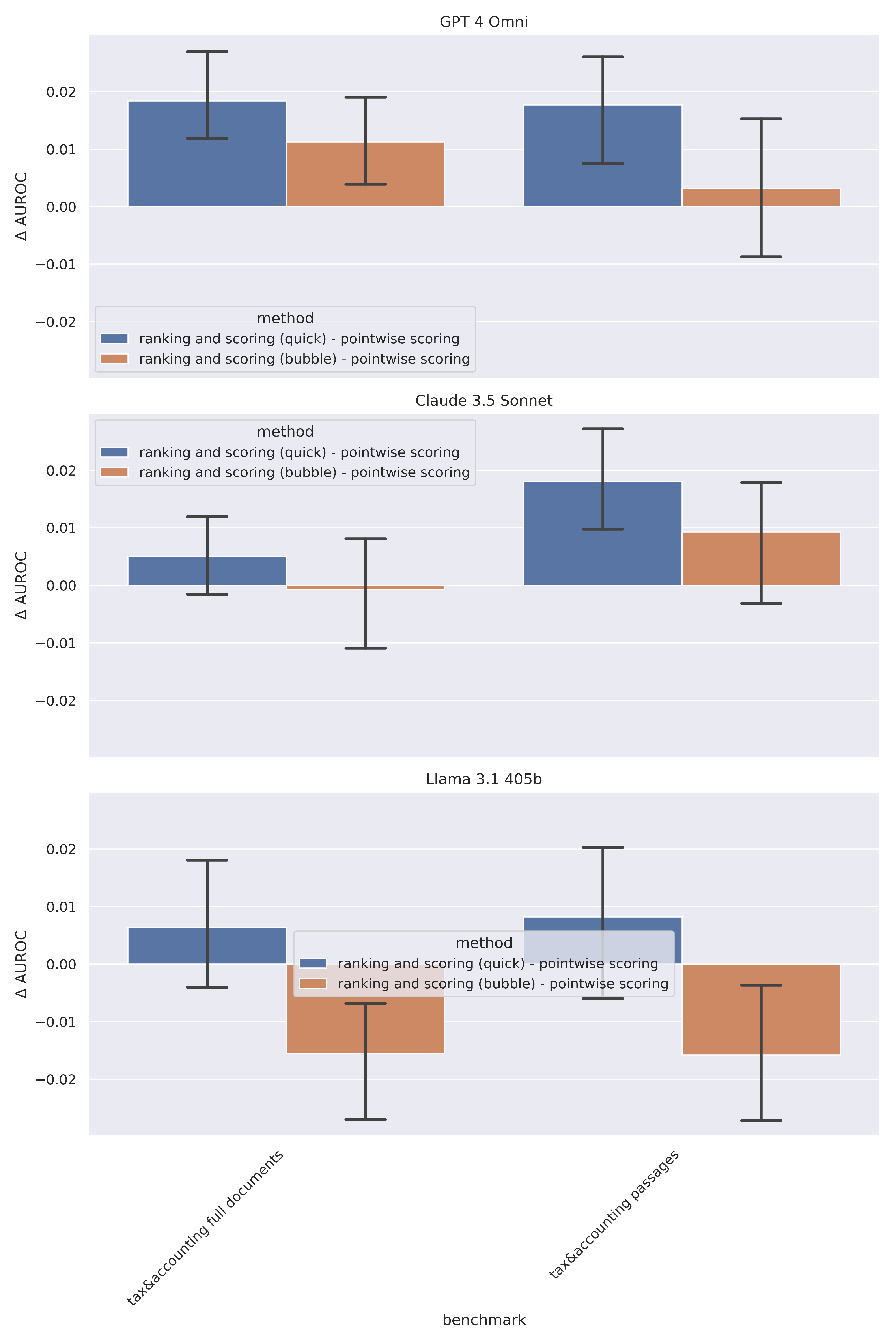}
        \caption{Difference in AUROC for pointwise scoring and listwise ranking on Tax\&Accounting datasets.}
    \end{subfigure}
    \hfill
    \begin{subfigure}[tb]{0.45\linewidth}
        \includegraphics[width=\linewidth]{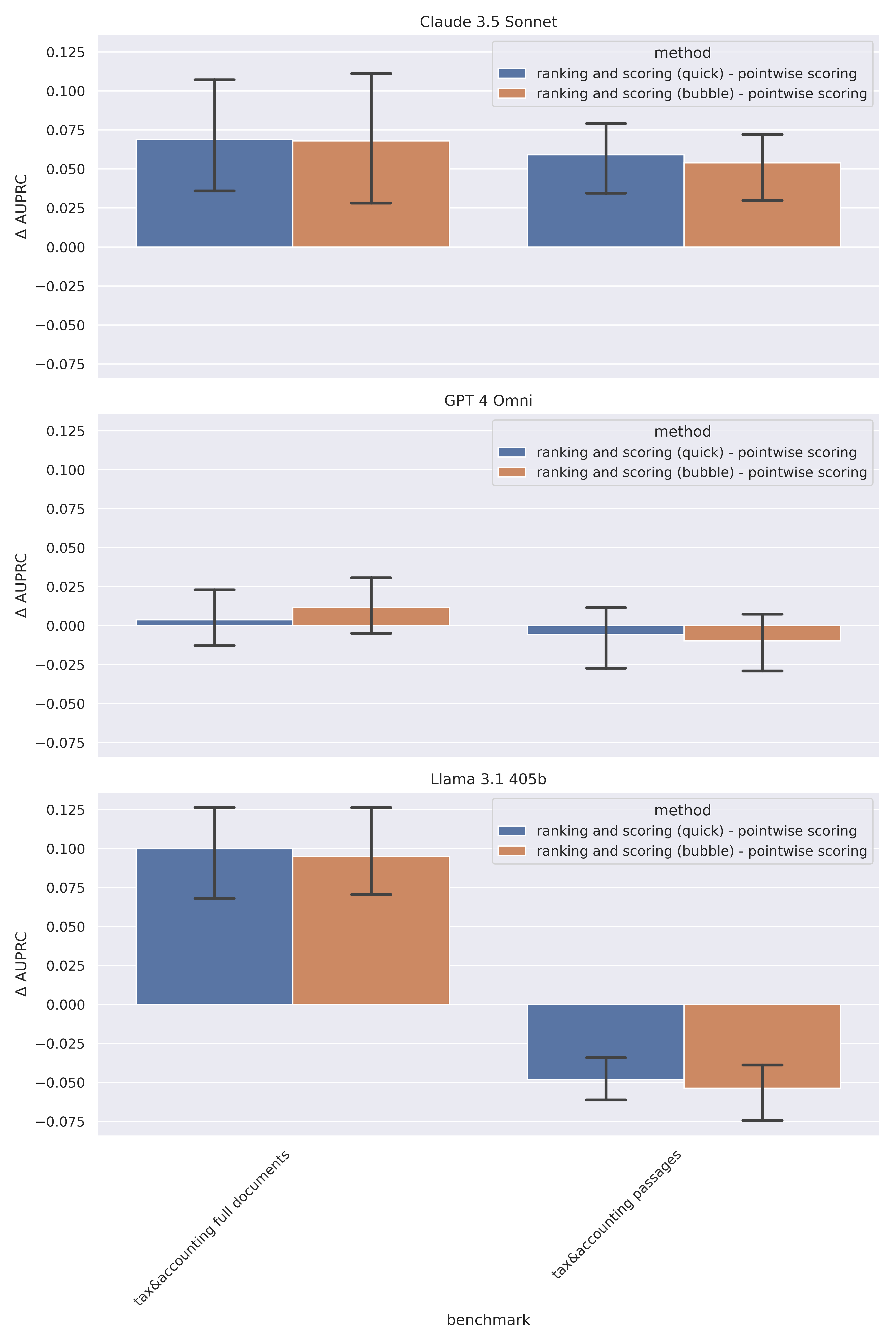}
        \caption{Difference in AUPRC for pointwise scoring and listwise ranking on Tax\&Accounting datasets.}
    \end{subfigure}
    \caption{Difference in area under curve metrics for pointwise scoring and listwise ranking on Tax\&Accounting datasets.}
\end{figure}

\FloatBarrier

\section{Creating the Tax\&Accounting datasets}
\label{sec:datasetconstruction}

To collect documents for relevance labeling (from which the passages are later
derived), we applied depth pooling, i.e. using an ensemble of five different retrieval
systems (four keyword search engines targeting different parts of the document
and its metadata and one vector search using embeddings from a transformer
model) and pooling their top hits. Due to practical constraints of the
annotation platform employed, we collected 20 relevance labels for each query,
prioritizing the top hits of each search engine. When the union of all top 5
hits had cardinality at least 20, we selected a random subset of 20 documents.
In cases where the union had cardinality less than 20, we added a random sample
of the union of all top 10 hits (and then, if necessary, the union of all top 20
hits) to reach 20 documents. We note that according to domain experts, most
queries in our dataset can be answered by one or at most a few documents, hence
the pooling thresholds and the 20 document limit are not expected to be a
significant limitation.

For document-level relevance labels we use a 4-point rubric, with relevance
scale ``Irrelevant'', ``Tangential'', ``Relevant but doesn't provide a full
answer'' and ``Relevant.'' Annotation was performed by subject matter experts
provided with a document containing domain-specific instructions regarding the
relevance levels as well as examples of query-document pairs at each level --
this documentation was itself written by subject matter experts, shared ahead of
the evaluation and accessible via hyperlink in the grading interface. Annotation
was completed during March 2024 (note that this is after the
pre-training data cut-offs of all LLMs evaluated except Claude 3.5 Sonnet
\cref{tab:llm_training_data}) and resulted in 4,189 labeled query-document
pairs.\footnote{A significant number of queries occur in multiple paraphrased variants, hence
    their corresponding pooled hits overlap.}

Passage level relevance labels were collected in a second round of annotation.
First, we filtered out all documents labeled ``Irrelevant'' in the document
level annotation. We then subdivide the remaining documents into semantically
meaningful sections, such that, according to experts, in most cases all passages
from a section share the same relevance label. Many short documents consist of a
single section, and for such documents we simply apply the document-level relevance
label to all passages. In the remaining cases, for each section, we uploaded the
first passage for subject matter expert annotation; the resulting relevance
label is then applied to all passages in its containing section. Based on
feedback from the document-level annotation, we use a 3-point rubric for passage
level relevance labels: ``Irrelevant'', ``Neutral'' and ``Relevant'' (hence the
``Tangential'' and ``Relevant but doesn't provide a full answer'' labels are
merged into ``Neutral''). Annotation was completed during May 2024 and resulted
in 11,534 labeled query-passage pairs.

\section{Prompt templates and API specifics}
\label{sec:prompttemplates}

For each of the three families of methods comprised of pointwise scoring, pure
permutation generation listwise ranking and simultaneous ranking+scoring, we
use a uniform system prompt across all LLMs.
\begin{lstlisting}[style=commoncode]
Pointwise Scoring System: You are an AI assistant tasked with evaluating a search result based on its relevance to a user's query. Your goal is to analyze the search result and assign it a relevance score.

Pure Permutation Generation System: You are an AI assistant tasked with ranking search results based on their relevance to a user's query.

Simultaneous Ranking and Scoring System: You are an AI assistant tasked with ranking and scoring search results based on their relevance to a user's query.
\end{lstlisting}
Our human prompt templates differ slightly from LLM to LLM, since we make a best
effort to follow their respective prompt engineering guidelines (for example,
the Claude documentation advises use of XML formatting, which that family if
models has allegedly been tuned to understand). For conciseness we focus mostly
on prompt templates for GPT 4 Omni.

For pointwise scoring, we use the following human message template:
\begin{lstlisting}[style=commoncode]
User query: """{user_query}"""
Search result:
"""
{search_result}
"""
Use the following 0-{max_points} scale to score the relevance of the search result:
{scale}
Instructions:
1. Carefully read and understand the content of the search result.
2. Compare it to the user's query, considering how well it addresses the user's information need.
3. Determine a relevance score based on the scoring system above.
Provide your score as a JSON dictionary with the following format:
```json
{{"score": integer in the range 0-{max_points} representing the relevance score of the search result}}
```
Reminder: the user's query is "{user_query}"
\end{lstlisting}
In this template, \texttt{user\_query} is a placeholder for the user query,
\texttt{search\_result} is a placeholder for the (first 300 words of the)
document to be scored, \texttt{max\_points} is a placeholder for the maximum
number of points on the relevance scale (1, 2, 4, 6 or 10), and \texttt{scale}
is a placeholder for the relevance scale itself. We enforce the schema of the
JSON output using OpenAI's structured output generation feature.

Our relevance scales are as follows:
\begin{lstlisting}[style=commoncode]
2-point scale: 
1: Relevant - The document addresses the user's query well, providing useful information related to the topic.
0: Not relevant - The document does not address the user's query or provides little to no useful information.
\begin{lstlisting}[style=commoncode]
3-point scale:
2: Excellent match, addresses the query comprehensively
1: Partial match, addresses some aspects of the query
0: Poor match, barely relevant or irrelevant to the query
\end{lstlisting}
\begin{lstlisting}[style=commoncode]
5-point scale:
4: Excellent match, addresses all or nearly all aspects of the query comprehensively
3: Good match, covers most key aspects of the query
2: Moderate match, partially relevant to the query
1: Poor match, only marginally related to the query
0: Irrelevant, no meaningful connection to the query
\end{lstlisting}
\begin{lstlisting}[style=commoncode]
7-point scale:
6: Perfect match, addresses all aspects of the query comprehensively
5: Excellent match, covers almost all aspects of the query in detail
4: Good match, addresses most aspects of the query
3: Average match, partially relevant to the query
2: Below average match, touches on the query topic but lacks depth
1: Poor match, only marginally related to the query
0: Completely irrelevant, no connection to the query
\end{lstlisting}
\begin{lstlisting}[style=commoncode]
11-point scale:
10: Perfect match, addresses all aspects of the query comprehensively
9: Excellent match, covers almost all aspects of the query in detail
8: Very good match, addresses most aspects of the query
7: Good match, covers several key aspects of the query
6: Above average match, addresses some important aspects of the query
5: Average match, partially relevant to the query
4: Below average match, touches on the query topic but lacks depth
3: Poor match, only marginally related to the query
2: Very poor match, barely relevant to the query
1: Extremely poor match, only contains a keyword or phrase from the query
0: Completely irrelevant, no connection to the query
\end{lstlisting}

For pure permutation generation, we use the following human message template:
\begin{lstlisting}[style=commoncode]
User query: """{user_query}"""
Search results:
"""
{search_results}
"""
The search results consist of a list of documents. Each document has the form:
<document document_id="unique identifier of this document">
<title>Title of This Document</title>
<content>Content of this document</content>
</document>
For each search result:
1. Carefully read and understand the content.
2. Compare it to the user's query, considering how well it addresses the user's information need.
Provide your response as a JSON object with a top level key "ranked_documents", whose value is a list of dictionaries where each dictionary contains the following key-value pairs:
- "document_id": unique identifier found as an attribute in the opening <document document_id="unique identifier of this document"> tag
- "rank": rank of the document based on relevance
Example of valid JSON format:
```json
{{"ranked_documents": [{{"document_id": "unique_id_1", "rank": 1}}, {{"document_id": "unique_id_2", "rank": 2}}, ...]}}
```
Ensure that the "ranked_documents" list is sorted in decreasing order of relevance, with the most relevant documents appearing first. The search results contain {num_documents} documents, so the document ranks should end at {num_documents}. Reminder: the user's query is """{user_query}"""
\end{lstlisting}
In this template \texttt{user\_query} is again placeholder for the user query, \texttt{search\_results} is a placeholder for the documents to be ranked, and \texttt{num\_documents} is a placeholder for the number of documents to be ranked (20 in our experiments). The prompt template suggests the template we use for the documents: we internally build a bijective map from its database unique identifier to an interger in the range 0-\texttt{num\_documents} (we term this an ``LLM friendly identifier''). Then for each document we build a string of the form
\begin{lstlisting}[style=commoncode]
<document document_id="{LLM friendly identifier}">
<title>document title</title>
<content>first 300 words of document</content>
</document>
\end{lstlisting}
These strings are then concatenated with newlines to form the \texttt{search\_results} variable. Again, the JSON output schema is enforced using OpenAI's structured output generation feature.

For simultaneous ranking+scoring, we use a hybrid template:
\begin{lstlisting}[style=commoncode]
User query: """{user_query}"""
Search results:
"""
{search_results}
"""
The search results consist of a list of documents. Each document has the form:
<document document_id="unique identifier of this document">
<title>Title of This Document</title>
<content>Content of this document</content>
</document>
Use the following 0-{max_points} scale to score the relevance of each search result:
{scale}
For each search result:
1. Carefully read and understand the content.
2. Compare it to the user's query, considering how well it addresses the user's information need.
3. Determine a relevance score based on the scoring system above.
Provide your response as a JSON object with a top level key "ranked_documents", whose value is a list of dictionaries where each dictionary contains the following key-value pairs:
- "document_id": unique identifier found as an attribute in the opening <document document_id="unique identifier of this document"> tag
- "rank": rank of the document based on relevance
- "score": relevance score of the document
Example of valid JSON format:
```json
{{"ranked_documents": [{{"document_id": "unique_id_1", "rank": 1, "score": 8}}, {{"document_id": "unique_id_2", "rank": 2, "score": 6}}, ...]}}
```
Ensure that the "ranked_documents" list is sorted in decreasing order of relevance, with the most relevant documents appearing first. The search results contain {num_documents} documents, so the document ranks should end at {num_documents}. Reminder: the user's query is """{user_query}"""
\end{lstlisting}
\emph{Note: the output format example above was used with the 11-point scale,
    for smaller scales one would need to ensure only valid relevance labels
    occur in the output example.}

\textbf{Prompt Engineering Process}: Prompts and relevance label rubrics were
generated using Anthropic's Claude Workbench ``generate a prompt'' tool once per
ordinal scale. Prompts were then adapted for each LLM based on their respective
documentation (e.g., ``end of turn'' symbols for Llama 3). For GPT 4 Omni and
Gemini 1.5 Pro, prompts were co-designed with their structured output features.
No modifications were made based on evaluation results, except for ensuring
Llama 3.1 405b adhered to output schemas by adding instructions like ``Provide
ONLY the JSON response in the output.''

\subsection{LLM APIs}

We use the following LLM APIs in our experiments. All evaluations were conducted
in late 2024/early 2025.

\begin{itemize}
    \item GPT 4 Omni: Microsoft Azure, via the official Python SDK, with the
          model identifier \texttt{gpt-4o}.
    \item Claude 3.5 Sonnet: Anthropic (first party), via the official Python
          SDK, with the model identifier \texttt{claude-3-5-sonnet-20240620}.
    \item Gemini 1.5 Pro: Vertex AI, via the official Python SDK, with the model identifier \texttt{gemini-1.5-pro}.
    \item Llama 3.1 405b: AWS Bedrock, via the official Boto3 Client, with the
          model identifier \texttt{meta.llama3-1-405b-instruct-v1:0}.
\end{itemize}

\section{Methods details}
\label{sec:methodsdetails}

\begin{algorithm}[tb]
    \caption{Bubble Sort Based Listwise Ranking With Telescoping}
    \label{alg:bubble_sort}
    \begin{algorithmic}
        \STATE {\bfseries Input:} query $q$, documents $D = d_1, \dots, d_N$,
        bootstrapping scores $b_1, \ldots, b_N$, window size $W$, overlap $V$, truncation thresholds $N=T_0, T_1, \ldots, T_m$
        \STATE \# Sort documents by decreasing bootstrapping score
        \STATE $D = \texttt{sorted}(D, \texttt{key} = b_i)$
        \STATE \# $p$ for ``pass''
        \FOR{$p=0$ {\bfseries to} $m$}
        \FOR{$i=T_p-W$ \textbf{to} $0$ \textbf{step} $W-V$}
        \STATE $window \leftarrow D[i:i+W]$
        \STATE $ranks, scores \leftarrow \texttt{Rank}(window)$
        \STATE \# Notation
        \STATE $\rho_i \leftarrow \text{rank of document} d_i$
        \STATE $r_i \leftarrow \text{relevance score of document} d_i$
        \STATE Update $d_i, d_{i+1}, \ldots, d_{i+W-1}$ with $scores$ (append $r_i$ to
        running list of relevance scores of document $d_i$)
        \STATE $D[i:i+W] \leftarrow \texttt{sorted}(D[i:i+W], \texttt{key}=\rho_i)$
        \ENDFOR
        \ENDFOR
        \STATE Average relevance scores for each document
        \STATE {\bfseries Output:} ranked and scored documents $D$
    \end{algorithmic}
\end{algorithm}

\begin{algorithm}[tb]
    \caption{Multi-Pivot Quicksort Based Listwise Ranking With Telescoping}
    \label{alg:quicksort}
    \begin{algorithmic}
        \STATE {\bfseries Input:} query $q$, documents $D = [d_1, \dots, d_N]$,
        bootstrapping scores $B = [b_1, \ldots, b_N]$, window size $W$, number of
        pivots $P$, truncation thresholds $N=T_0,
            T_1, \ldots, T_m$
        \STATE \# $p$ for ``pass''
        \FOR{$p=0$ {\bfseries to} $m$}
        \STATE \# Select $P$ pivots. Here \texttt{SelectPivots} encapsulates uniform sampling on the distribution of bootstrapping scores.
        \STATE $pivots \leftarrow \texttt{SelectPivots}(D[0:T_p], \texttt{bootstrapping\_scores}=B[0:T_p], \texttt{num\_pivots}=P)$
        \STATE $batches \leftarrow \texttt{DivideIntoBatches}(D[0:T_p] \setminus pivots, W-P)$
        \FOR{each batch in $batches$}
        \STATE $ranks, scores \leftarrow \texttt{Rank}(pivots \cup batch)$
        \STATE Update documents in $pivots \cup batch$ with $ranks, scores$, and
        ``pivots to left and right (see text)'' (append to appropriate running lists)
        \ENDFOR
        \STATE \# Notation
        \STATE $\bar{\rho_j} \leftarrow \text{average rank of document} d_j$
        \STATE $\bar{r_j} \leftarrow \text{average relevance score of document}
            d_j$
        \STATE \# Compute ``pivot scores''
        \STATE $s_j \leftarrow (-\bar{\rho}_j, \bar{r}_j)$ \# just for documents
        $p_j$ in $pivots$
        \STATE \# Extend pivot scores to non-pivot documents
        \STATE $p_{L(i)}, p_{R(i)} \leftarrow \text{pivots to left and right of
            } d_i$ \# for documents $d_i \in D[0:T_p] \setminus pivots$,
        \STATE $s_i \leftarrow \frac{1}{2}(s_{p_L(i)} + s_{p_R(i)})$ \# for
        documents $d_i \in D[0:T_p] \setminus pivots$
        \IF{ranking+scoring}
        \STATE \# replace bootstrapping scores $b_i$ with LLM relevance
        scores $r_i$ (or their averages in the case of pivots)
        \STATE $b_i \leftarrow \bar{r}_i$ \# for alldocuments $d_i \in D[0:T_p]$
        \ENDIF
        \STATE \# Lexicographic sorting: sort documents by decreasing $s_i$, then decreasing $r_i$,
        then decreasing $b_i$ (using python \texttt{sorted} conventions)
        \STATE $D[0:T_p] = \texttt{sorted}(D[0:T_p], \texttt{key} = (s_i, r_i, b_i),
            \texttt{reverse}=\texttt{True})$
        \ENDFOR
        \STATE {\bfseries Output:} ranked and scored documents $d_i$
    \end{algorithmic}
\end{algorithm}

As mentioned in \cref{sec:implementations}, we consider two methods (bubblesort
and quicksort) for bootstrapping a listwise (scorer and) ranker capable of
handling, say, 20 documents at a time to an algorithm that can rank and score
lists of arbitrary size.

Both of these leverage what we call a "bootstrapping score" $b_i$ that coarsely
sorts documents by relevance (in prior work and our experiments, this is a score from an
upstream retrieval system like BM25 or embedder cosine similarity).

The first is bubble sort \cite{sunChatGPTGoodSearch2023}. This sorting algorithm is well
adapted to ranking and ranking+scoring for two reasons:
\begin{itemize}
    \item It is easy to generalize the standard bubble sort algorithm to make it
          leverage $n$-ary comparisons with $n > 2$ (using sliding windows of size $n$).
    \item While the average and worst case sorting performance of this algorithm are
          $O(N^2)$, the performance on lists that are "almost sorted" is roughly
          $O(N)$. This is
          the expected scenario when ordering documents by decreasing bootstrapping score
          prior to bubble sort.
    \item It is straightforward to "early stop" this algorithm (simply by limiting
          the number of bubbling-up passes through the list).
\end{itemize}
When simultaneously ranking+scoring, documents will be scored multiple times
during bubble sort (due to overlap of sliding windows). To account for this, we
maintain lists of the scores each document has received during the algorithm.
Once the bubble sort algorithm terminates, for each document we replace that list
with its average, rounded to the nearest integer relevance score. In all of our
experiments, we use a sliding window size of 20 and an overlap of 10. For
further details on this method, we refer the reader to
\cite{sunChatGPTGoodSearch2023,maZeroShotListwiseDocument2023} (though we
implement with our own code we follow their implementations closely). See \cref{alg:bubble_sort} for a pseudocode description of the algorithm.

The second algorithm we consider is a variant of quicksort
\cite{parryTopPartitioningEfficientListWise2024,qinLargeLanguageModels2023a}.
The first modification we make is replacing binary comparisons with $n$-ary
comparisons and using $1 < k < n$ pivot points. Our motivations for this change
include:
\begin{itemize}
    \item In the case where relevance comparisons are transitive, and where one
          runs quicksort to completion, the number of pivots used does not affect the
          final ranking. \emph{But} increasing the number of pivots can improve the
          coarse ranking of the list after, say, the first quicksort pass. Consider a
          thought experiment where we are ranking 100 documents. If we use a single
          pivot which happens to be the 50th most relevant document, then after the
          first quicksort pass, we will have identified the top and bottom 50
          documents, but (based only on comparisons with the pivot) with no
          information about relevance ranking within the top and bottom halves. If we
          use 4 pivots, which happen to the 20th, 40th, 60th and 80th most relevant
          documents, then after the first quicksort pass we will have identified the
          top 20, 20-40, and so on documents. Even though we may not have identified
          the relevance ranking within the top twenty documents, for
          precision-oriented tasks like top-k ranking, this is a significant
          improvement.
    \item \citet{parryTopPartitioningEfficientListWise2024} points out that
          their ``top down partitioning'' algorithm, which can be viewed as a sort of
          greedy quicksort that picks a single pivot and, beginning at the top of the
          coarsely sorted list, looks for candidate documents more relevant than the
          pivot, is \emph{sensitive to the choice of pivot}. We hypothesize that using
          more than one pivot, we hedge against this sensitivity.
    \item Seting the number of quicksort pivots equal to the overlap of bubblesort sliding windows
          has the arguably appealing effect of making the computational cost of
          our bubblesort and quicksort variants essentially equal.
\end{itemize}
We sample pivots $p_1, \ldots, p_k$ uniformly according to their bootstrapping
score.  We then randomly divide the non-pivot documents into batches $B_1,
    \ldots, B_m$ of size  at most $n-k$, and rank and score them together with
the pivot documents. In all of our experiments we use \(n=20\) and \(k=10\).

For brevity in what follows we describe the quicksort extension of \emph{scoring
    and ranking}. For the case of pure permutation generation ranking, the
scores are simply set to a default value of 0; since they are only ever used
for tie-breaking in ranking+scoring, this has no adverse effect.

For non-pivot documents $d_i$, which are scored and ranked exactly once per
quicksort pass, we keep track of their score $r_i$ as well as the pivot
documents $p_{L(i)}$ and $p_{R(i)}$ on their left and right in the sorted
list. For pivot documents $p_j$, which are scored and ranked $m$ times, we
accumulate a list of scores $R_j = (r_{j_1}, \ldots, r_{j_m})$ and a list of
ranks $\rho_j = (\rho_{j_1}, \ldots, \rho_{j_m})$. For technical convenience
we also declare a pivot to be its own "pivot to the left and right", i.e.
$p_L(j) = p_R(j) = p_j$. As has been observed elsewhere many times, a
nuisance endemic to LLM ranking is that there are no guarantees that
relative rankings are transitive. To robustly order the list after ranking
all batches we proceed as follows:
\begin{itemize}
    \item Each pivot document $p_j$ is assigned a temporary "pivot
          score" $s_j = (-\bar{\rho}_j, \bar{r}_j)$, where $\bar{\rho}_j$ and $\bar{r}_j$
          are the average rank and score. Thus a pivot gets a high score if it
          typically gets ranked in the top of its batch, and if ties occur, we
          break them using average relevance score.
    \item We then extend "pivot scores" to the entire list of documents by
          averaging the "pivot-scores" of the document's left and/or right
          pivots: for all documents $d_i$, we set $s_i = \frac{1}{2}(s_{p_L(i)}
              + s_{p_R(i)})$. Thus a non-pivot document acquires a high score if it typically appears between highly ranked pivots.
    \item We then sort the entire list of documents lexicographically
          (\emph{decreasing}) on: $s_i$, $r_i$, $b_i$. Thus the dominant term in
          the sort depends on comparison against pivots and ties are broken
          using the LLM relevance score, then bootstrapping score.
\end{itemize}
See \cref{alg:quicksort} for a pseudocode description of the algorithm.

Both bubble and quicksort have worst case time complexity $O(N^2)$ where $N$ is
the number of documents to be ranked. As in prior work (see
\cref{sec:related-work}), to obtain time complexity linear in the number of
documents to be ranked, we use early-stopping versions of these algorithms.
While taking the classical sorting algorithm analogy verbatim would suggest
that in the case of bubblesort we fix a number of bubbling up passes, and in
the case of quicksort we fix a maximum recursion depth, in our experiments we
instead use a simple ``telescoping'' procedure as follows: for a list of size $N$ (and in
our main experiments, $N=100$) we fix truncation thresholds $N > T_1 > T_2 >
    \ldots > T_m > 1$ (in our main experiments, $m=2$, $T_1 = 50$ and $T_2=20$). We then
run one pass (in the case of bubble sort, only a single bubbling up pass and in the
case of quicksort, a single pass through the list comparing all documents with
pivots but no recursion) on the size $N$ list, then truncate to the top
$T_1$ documents, run once more, truncate to the top $T_2$ documents, and so on.

\textbf{Note}: In the case of quicksort, after the first pass, we replace the
bootstrapping score $b_i$ (initially equal to BM25 score) with the LLM score
$r_i$. The motivation for this is that we hypothesize (but do not attempt to
demonstrate empirically or theoretically) that multi pivot quicksort
works best when the pivots have a uniformly distributed range of relevance,
since in that case after comparing all documents to pivots and partitioning the
list accordingly, the max size of a partition is minimized.\footnote{An analogy
    to classical binary comparison quicksort, this is the scenario when the pivot
    happens to be the midpoint of the list.} Since the hypothesis is that LLM
relevance scores are better than those of BM25, we attempt to use them to
improve sampling of pivots.

The motivation for telescoping is a focus on information retrieval metrics of
the form ``X@'' for relatively small $K$. We seek to iteratively refine shorter
and shorter truncations of the ranked document list. In the case of bubble sort,
this also can be viewed as a prior that the coarse ranking by bootstrapping
score is good. Technically speaking, with the parameters of our experiments the
telescoping procedure implies that we can promote any 10 of the 100 documents to
the top 10, and any 20 of the top 50 documents to the top 20. In the case of
quicksort, telescoping is conceptually similar to the ``candidate selection with
a budget'' approach of \cite{parryTopPartitioningEfficientListWise2024}, where
documents ranked below a pivot are not considered for later ranking refinement
passes (though we cut off at a fixed number of documents, not a pivot). It is
also conceptually similar to restricting to the top ranked partition(s) at every
recursive call of quicksort. Note however that these iterative refinement
methods based on cutting off at  pivots have no obvious analog for bubble sort,
whereas the telescoping procedure can be applied to both algorithms, which we
found convenient.

A final note on telescoping: multiple of our references in
\cref{sec:related-work} consider only one pass of bubblesort or quicksort. We
chose to add iterative refinement since our core claim is that pointwise scoring
yields information retrieval performance comparable to listwise ranking on many
LLM-benchmark dataset combinations, and as such we wanted to err on the side of
over-optimizing listwise ranking.

\subsection{Handling LLM inference errors}
\label{sec:handlingllminferenceerrors}

Our experiments consume many thousands of LLM API calls. For any given API call
ranking and or scoring one or more documents comma, there are several possible
causes of error, for example a network error, a timeout, or an LLM output that
does not conform to the expected output schema.\footnote{Structured output
    generation capabilities, for the APIs that had them at the time of running
    experements, drop this error rate to effectively zero.} We observe the latter more
frequently for listwise ranking than for pointwise scoring, since the latter
requires only a single integer output per LLM call. A closely related error we
observed for listwise ranking and ranking+scoring (also observed in prior work) is
failure to return all document IDs.

Our approach to handling these errors is twofold:

\textbf{Retries}: Our experimental code includes several layers of retries. At
the lowest level, all of LLM our per-query-document pointwise scoring inferences
and batch-of-documents listwise ranking inferences are retried up to 3 times,
with a 2 second delay between retries. Here ``inferences'' includes both the LLM
call itself as well as the postprocessing of the LLM output according to the
specified schema.

\textbf{Fallback values}:
In the case of pointwise scoring, if after 3 retries we do not get a valid LLM
relevance score, we assign a relevant score of zero. In the case of listwise
ranking, if one of the documents in a batch does not get a valid LLM relevance
rank and/or score, we assign it a rank of $N$ (the number of documents in the
batch) and a relevance score of zero.

In the event that we run through all retries and fail to rank documents for a
given query, then that query is omitted from the resulting \texttt{trec\_eval}
run file. To the authors best understanding the default behavior of
\texttt{trec\_eval} is to assign a query-level score of zero in this case (for
all metrics we consider this reasonably encapsulates ``the system crashed and
returned no relevant documents'').

\section{Efficiency considerations}
\label{sec:efficiencyconsiderations}

While the purpose of this paper is not to compare the efficiency of the
different LLM relevance ranking methods studied (but rather their performance),
in this section we make a few comments on the topic.

\subsection{Token usage}

\begin{table}[ht]
    \centering
    \caption{Approximate token usage for different reranking methods,
        where $N$ is the number of documents, $L$ is the average number of tokens per document,
        $W$ is the window size, $V$ is the sliding window overlap,
        $P$ is the number of pivots and \( N > T_1 > T_2 > \dots \) are the
        truncations used for telescoping. Here \(c\) is a small constant
        accounting for the number of output tokens consumed per document
        (document ID, document rank, relevance label, etc.).}
    \begin{tabular}{lcc}
        \toprule
        \textbf{Method}     & \textbf{Input tokens}                                                    & \textbf{Output tokens}                                           \\
        \midrule
        Pointwise scoring   & $N \cdot L$                                                              & $N$                                                              \\
        Bubble sort         & $(1 + \sum_{i} \frac{T_i-V}{N-V}) \cdot \frac{N-V}{W -
        V} \cdot W \cdot L$ & $(1 + \sum_{i} \frac{T_i-V}{N-V}) \cdot \frac{N-V}{W - V} \cdot cW$                                                                         \\
        Quicksort           & $(1 + \sum_{i} \frac{T_i-P}{N-P}) \cdot \frac{N-P}{W-P} \cdot W \cdot L$ & $(1 + \sum_{i}\frac{T_i-P}{N-P}) \cdot \frac{N-P}{W-P} \cdot cW$ \\
        \bottomrule
    \end{tabular}
\end{table}

Consider a fixed query and list of documents to be ranked by relevance with
respect to the query. Let \(N\) be the length of the list of documents to be
ranked, and let \(L\) be the average number of tokens per document. In our
experiments, \(N=100\) and \(L \approx 500\) (see below for real token
statistics).  Throughout this discussion, we'll make this simplifying assumption
that the number of input tokens consumed by the prompt template and query are
insignificant compared to the number of input tokens consumed by any one
document in the list of documents to be ranked. A few observations:

For pointwise scoring, the number of input tokens consumed when ranking the list
of \(N\) documents is \(\approx N \cdot L\), and the number of output tokens is
roughly \(N\) (essentially one integer in the range \(0-10\) per document).

For bubble sort based listwise ranking, let \(W\) be the sliding window size and
let \(V\) be the size of the sliding window overlap (in our experiments \(W =
20, V=10\), following \cite{sunChatGPTGoodSearch2023}). Observe that for each window of \(W\)
documents ranked we consume \(W \cdot L\) input tokens and produce on the order
of \(cW\) output tokens where \(c\) is a small integer constant (for each of the
\(W\) documents we get a document ID, a document rank and possibly also a
document score, all integers in the range 0-20). The number of windows is
roughly \(\frac{N-V}{W-V}\), hence the total number of input tokens consumed is
roughly \(\frac{N-V}{W-V} \cdot W \cdot L\) and the total number of output tokens
is roughly \(\frac{N-V}{W-V}\cdot cW\).

Coming up with an estimate for quicksort based listwise ranking is similar. Let
\(W \) be the total batch size (number of documents sent to the LLM at one time,
including pivots), and let \(P\) be the number of pivots (so \(P < W\)). In our
experiments \(W=20\) and \(P=10\). Again
for each batch of \(W\) documents ranked we consume \(W \cdot L\) input tokens
and produce on the order of \(cW\) output tokens. The number of batches is
roughly \(\frac{N-P}{W-P}\), hence the total number of input tokens consumed is
\(\frac{N-P}{W-P} \cdot W \cdot L\) and the total number of output tokens is
roughly \(\frac{N-P}{W-P} \cdot cW\).

Recall that for bubble and quicksort based listwise ranking we use a
``telescoping'' procedure, where we first rank the top \(N\) documents, then
truncate to the top \(T_1\) documents, then rank those, truncate to the top
\(T_2\) documents, and so on. This means that the total number of input tokens
will be scalled by the number of, say, ``effective passes'' through the list of
documents. In closed form, for bubble sort based listwise ranking we end up with
\[ \text{input tokens} \approx (1 + \sum_{i} \frac{T_i - V}{N -V}) \cdot \frac{N-V}{W-V}
    \cdot W \cdot L, \text{  and output tokens} \approx (1 + \sum_{i} \frac{T_i-V}{N-V})
    \cdot \frac{N-V}{W-V} \cdot cW. \]
Similarly, for quicksort based listwise ranking we have
\[ \text{input tokens} \approx (1 + \sum_{i} \frac{T_i-P}{N-P}) \cdot \frac{N-P}{W-P}
    \cdot W \cdot L, \text{  and output tokens} \approx (1 + \sum_{i} \frac{T_i-P}{N-P}) \cdot \frac{N-P}{W-P} \cdot cW. \]

In our experiments, we use \(T_1=50\) and \(T_2=20\) hence the multiplier \( 1 +
\sum_{i} \frac{T_i}{N} = 1 + \frac{50}{100} + \frac{20}{100} = 1.7\).

Plugging in all values used in our experiments, we find that as implemented in
this paper the listwise ranking methods consume roughly \(2 \cdot 1.7 = 3.4\)
times the number of input and output tokens consumed by pointwise scoring.

\subsection{Parallelism, Batching and API calls}
\label{sec:parallelism-batching}

While the previous section shows that the three different methods under
discussion use the same number of input/output tokens up to an order of
magnitude, they have quite different parallelism properties.

Pointwise scoring is ``embarrassingly parallel,'' in that the relevance of
each document can be computed independently of the others. The parallelism degree
(maximum number of documents being ranked at any given instant) is \(N\), and in the absence of any
batching mechanism\footnote{Numerous LLM providers are rolling out ``batch
    APIs,'' however in some cases these seem to be designed for offline processing jobs
    hence have unclear latency properties.} this coincides with the number of API
calls required.

Bubble sort based listwise ranking lies at the opposite end of the spectrum: the
\(\approx  \frac{N-V}{W-V}\) sliding windows must be processed in order during
a bubbling up pass, hence the parallelism degree is \( 1\). This implies \(\approx  \frac{N-V}{W-V}\) API calls (or
\(\approx (1 + \sum_{i} \frac{T_i-V}{N-V}) \cdot \frac{N-V}{W - V} \) with telescoping), which
must be run in serial.

Quicksort based listwise ranking lies somewhere in between. The
\(\frac{N-P}{W-P}\) batches of documents to be ranked alongside the pivots can
be processed in parallel (previously pointed out in
\citet{qinLargeLanguageModels2023a,parryTopPartitioningEfficientListWise2024}),
hence the parallelism degree is \(\approx \frac{N-P}{W-P}\). In the case of
telescoping, after the first truncation the parallelism degree becomes \(\approx
\frac{T_1-P}{W-P}\), after the second truncation \(\approx \frac{T_2-P}{W-P}\),
and so on.

Note that many LLM providers place limits on the number of concurrent API
calls. With the parameters of our experiments (\(N=100, W=20, V=P=10\)), and
assuming a maximum concurrent API call limit of \(10\),\footnote{and assuming that the
    pointwise scoring inputs cannot be directly batched into API calls or directly
    batched into an LLM forward pass if one is hosting the model.} the effective parallelism
degree of quicksort based listwise ranking, which is \(\approx 9\) up to the
first truncation (then 4, then 1), comes close to that of pointwise scoring
which is \(\approx 10\) in this thought experiment.

\begin{table}[ht]
    \centering
    \caption{Parallelism and API call estimates for different ranking methods,
        where $N$ is the number of documents, $W$ is the window size, $V$ is the sliding window
        overlap, $P$ is the number of pivots.}
    \begin{tabular}{lcc}
        \toprule
        \textbf{Method} & \textbf{Max Parallelism}     & \textbf{\# of API Calls}                 \\
        \midrule
        Pointwise       & $N$                          & $N$                                      \\
        Bubble Sort     & $1$                          & $\approx \frac{N - V}{W - V}$ (per pass) \\
        Quicksort       & $\approx\frac{N - P}{W - P}$ & $\approx\frac{N - P}{W - P}$ (per level) \\
        \bottomrule
    \end{tabular}
\end{table}